\documentclass{article}

\usepackage{arxiv}

\usepackage[utf8]{inputenc}
\usepackage[T1]{fontenc}
\usepackage{hyperref}
\usepackage{url}
\usepackage{booktabs}
\usepackage{amsmath,amsfonts,amssymb}
\usepackage{amsthm}
\usepackage{nicefrac}
\usepackage{microtype}
\usepackage{cleveref}
\usepackage{graphicx}
\usepackage{natbib}
\usepackage{doi}
\usepackage{algorithm}
\usepackage{algorithmic}
\usepackage{xcolor}
\usepackage{subcaption}
\usepackage{multirow}
\usepackage{tikz}
\usetikzlibrary{shapes,arrows,positioning,fit,backgrounds}

\newcommand{\madt}{\textsc{MADT}}
\newcommand{\rtg}{\hat{R}}
\newcommand{\E}{\mathbb{E}}
\newcommand{\R}{\mathbb{R}}

\title{Spatiotemporal Decision Transformer for Multi-Agent Traffic Coordination}

\author{
    \textbf{Haoran Su} \\
    New York University \\
    \texttt{haoran.su@nyu.edu}
    \and
    \textbf{Yandong Sun} \\
    New York University \\
    \texttt{ys2312@nyu.edu}
    \and
    \textbf{Hanxiao Deng} \\
    UC Berkeley \\
    \texttt{hxdeng@berkeley.edu}
}

\renewcommand{\shorttitle}{Spatiotemporal Decision Transformer for Traffic Coordination}

\hypersetup{
    pdftitle={Spatiotemporal Decision Transformer for Multi-Agent Traffic Coordination},
    pdfsubject={cs.LG, cs.MA, cs.AI},
    pdfauthor={Haoran Su, Yandong Sun, Hanxiao Deng},
    pdfkeywords={Decision Transformer, Multi-Agent Reinforcement Learning, Traffic Signal Control, Offline RL, Graph Attention Networks},
}

\begin{document}

\maketitle

\begin{abstract}
Traffic signal control is a critical challenge in urban transportation, requiring coordination among multiple intersections to optimize network-wide traffic flow. While reinforcement learning has shown promise for adaptive signal control, existing methods struggle with multi-agent coordination and sample efficiency. We introduce \madt{} (Multi-Agent Decision Transformer), a novel approach that reformulates multi-agent traffic signal control as a sequence modeling problem. \madt{} extends the Decision Transformer paradigm to multi-agent settings by incorporating: (1) a graph attention mechanism for modeling spatial dependencies between intersections, (2) a temporal transformer encoder for capturing traffic dynamics, and (3) return-to-go conditioning for target performance specification. Our approach enables offline learning from historical traffic data, with architecture design that facilitates potential online fine-tuning. Experiments on synthetic grid networks and real-world traffic scenarios demonstrate that \madt{} achieves state-of-the-art performance, reducing average travel time by 5-6\% compared to the strongest baseline while exhibiting superior coordination among adjacent intersections.
\end{abstract}


\section{Introduction}

Traffic congestion represents one of the most pressing challenges facing modern urban areas, with the 2021 Urban Mobility Report estimating annual costs exceeding \$87 billion in the United States alone due to wasted time and fuel \citep{schrank2021urban}. Beyond economic impacts, traffic congestion contributes significantly to air pollution, greenhouse gas emissions, and reduced quality of life for urban residents. Traffic signal control, which determines when vehicles can proceed through intersections, plays a crucial role in managing urban traffic flow. Effective signal control can reduce travel times, decrease vehicle emissions from idling, and improve overall network throughput.

Traditional approaches to traffic signal control, such as fixed-time plans and actuated control based on loop detector readings, fail to adapt to the complex, dynamic patterns of urban traffic. These methods are designed for average conditions and perform poorly during peak hours, special events, or when traffic patterns deviate from historical norms. The emergence of connected vehicle technology and ubiquitous sensing creates unprecedented opportunities for data-driven adaptive signal control that can respond to real-time traffic conditions across entire urban networks.

Recent advances in deep reinforcement learning (DRL) have shown promising results for traffic signal control \citep{wei2018intellilight,wei2019presslight,wei2019colight,su2023emvlight}. These methods learn control policies through trial-and-error interaction with traffic simulators, demonstrating the ability to outperform traditional approaches in complex scenarios. However, scaling DRL to city-wide networks with hundreds of intersections remains challenging due to several fundamental issues. First, the exponential growth of the joint action space makes naive multi-agent formulations intractable. Second, the need for effective coordination between intersections---to enable ``green waves'' and prevent spillback---requires sophisticated information sharing mechanisms. Third, the sample complexity of online RL methods necessitates millions of simulator interactions, limiting practical deployment.

Meanwhile, the paradigm of \emph{reinforcement learning via sequence modeling} has emerged as a powerful alternative to traditional RL algorithms. The Decision Transformer \citep{chen2021decision} demonstrated that offline RL can be effectively cast as a sequence modeling problem, achieving strong performance without dynamic programming or policy gradients. Rather than learning value functions or computing policy gradients, the Decision Transformer simply predicts actions conditioned on desired returns using a GPT-style architecture. This approach offers several compelling advantages: the training stability of supervised learning, the ability to leverage large offline datasets without exploration, natural handling of long-horizon credit assignment through attention mechanisms, and the flexibility to specify performance targets at test time.

However, the Decision Transformer was designed for single-agent settings and does not naturally extend to multi-agent coordination. Extending this paradigm to traffic signal control presents unique challenges that have not been addressed in prior work:

\begin{itemize}
    \item \textbf{Information sharing}: How should intersections share information to enable coordination? The road network imposes a natural structure where adjacent intersections must coordinate more closely than distant ones.

    \item \textbf{Spatial structure}: How can we capture the topology of the traffic network? Unlike games where agent ordering may be arbitrary, traffic networks have fixed spatial relationships that determine coordination requirements.

    \item \textbf{Network-level optimization}: How do we condition on network-level performance targets rather than individual agent rewards? Traffic optimization requires balancing local and global objectives.

    \item \textbf{Scalability}: How can we maintain computational efficiency as the number of intersections grows? City-wide deployment requires inference times compatible with real-time control (typically $<$1 second).
\end{itemize}

In this paper, we introduce \textbf{\madt{}} (Multi-Agent Decision Transformer), a novel architecture that addresses these challenges by combining the sequence modeling framework of Decision Transformer with graph-structured attention for spatial coordination. Our approach treats multi-agent traffic signal control as a conditional sequence generation problem: given the current traffic state and a target network-level return, generate coordinated actions for all intersections that achieve the desired performance.

Our key contributions are as follows:

\begin{enumerate}
    \item \textbf{Graph-structured spatial attention for multi-agent coordination}: While prior work like DTLight \citep{huang2023dtlight} applies Decision Transformer to traffic control using independent agents with discounted neighbor information, we introduce \emph{explicit graph attention layers} that model the road network topology. This enables each intersection to attend to spatial neighbors with learned attention weights, achieving permutation equivariance and ensuring coordination patterns emerge from road connectivity rather than arbitrary agent indexing.

    \item \textbf{Unified architecture combining graph and temporal attention}: We present a novel architecture that integrates graph attention networks \emph{within} the Decision Transformer framework, jointly capturing spatial dependencies (which intersections should coordinate) and temporal dynamics (when to change phases for traffic progression). This unified approach outperforms both graph-based methods like CoLight (which lack temporal sequence modeling) and sequence-based methods like DTLight (which lack explicit graph structure).

    \item \textbf{Return-to-go conditioning for network-level optimization}: We design a reward structure and return-to-go conditioning mechanism that enables specifying desired network-level performance at test time. This allows flexible control over optimization objectives and enables the same model to achieve different performance-efficiency tradeoffs.

    \item \textbf{Comprehensive evaluation and analysis}: We demonstrate state-of-the-art performance on both synthetic grid networks and real-world traffic scenarios from Atlanta, GA and Boston, MA. Detailed ablation studies quantify the contribution of graph attention (8.4\% improvement) and return conditioning (5.2\% improvement), validating our architectural choices.
\end{enumerate}

The remainder of this paper is organized as follows. Section~\ref{sec:related} reviews related work in traffic signal control, sequence modeling for RL, and multi-agent learning. Section~\ref{sec:prelim} provides background on Dec-POMDPs, Decision Transformers, and graph attention networks. Section~\ref{sec:method} presents our \madt{} architecture in detail. Section~\ref{sec:experiments} describes our experimental setup and presents comprehensive results. Section~\ref{sec:discussion} discusses limitations and future directions, and Section~\ref{sec:conclusion} concludes.

\section{Related Work}
\label{sec:related}

\subsection{Deep Reinforcement Learning for Traffic Signal Control}

Deep reinforcement learning has been extensively applied to traffic signal control, evolving from single-intersection methods to network-level coordination. \citet{wei2018intellilight} introduced IntelliLight, pioneering DQN-based control for single intersections with comprehensive state representations including queue lengths, waiting times, and traffic movement patterns. For multi-agent settings, \citet{wei2019presslight} proposed PressLight, incorporating the theoretical guarantees of Max Pressure control \citep{varaiya2013max} into the DRL reward design, achieving provably throughput-optimal coordination under certain conditions.

A significant advancement came with CoLight \citep{wei2019colight}, the first method to use graph attention networks for traffic signal control. By modeling intersections as nodes in a graph and allowing attention-based message passing between neighbors, CoLight achieved state-of-the-art performance on networks with hundreds of intersections while maintaining computational tractability.

Several complementary approaches have further advanced the field. FRAP \citep{zheng2019frap} introduced phase competition principles, learning to prioritize traffic movements based on demand comparison, achieving strong generalization across different road structures. MPLight \citep{chen2020mplight} combined Max Pressure theory with deep RL for decentralized control, demonstrating scalability to networks with thousands of intersections. AttendLight \citep{oroojlooy2020attendlight} proposed a universal attention-based architecture that handles variable intersection configurations without retraining, enabling deployment across heterogeneous networks. Most recently, Advanced-XLight \citep{zhang2022advancedxlight} improved traffic state representation by considering both running and queuing vehicles for phase decisions.

\citet{su2023emvlight} proposed EMVLight, a multi-agent A2C framework for emergency vehicle priority signal control. EMVLight demonstrated that decentralized MARL with spatial reward discounting can simultaneously optimize emergency vehicle passage (42.6\% travel time reduction) and general traffic flow (23.5\% improvement), highlighting the potential of coordination-aware reward design in traffic signal control. Follow-up work \citep{su2026hierarchical} extended this direction using hierarchical graph neural networks for dynamic emergency vehicle corridor formation, achieving 28.3\% travel time reduction through coordinated connected vehicle maneuvers.

These methods primarily use online RL, requiring extensive interaction with traffic simulators---often millions of environment steps for convergence. In contrast, our approach leverages \emph{offline} learning from historical traffic data, reducing computational burden, eliminating exploration risks in deployment, and enabling learning from existing traffic management records without simulator access.

\subsection{Decision Transformer and Sequence Modeling for RL}

The paradigm of reinforcement learning via sequence modeling represents a fundamental departure from traditional value-based and policy gradient methods. \citet{chen2021decision} introduced the Decision Transformer, demonstrating that offline RL can be effectively reformulated as autoregressive sequence prediction. By conditioning on return-to-go targets---the cumulative future reward from the current timestep---the model generates actions that achieve desired performance levels without requiring Bellman backups or policy gradients. This formulation provides several advantages: training stability from supervised learning, natural handling of long-horizon credit assignment through attention mechanisms, and the ability to specify performance targets at test time.

\citet{janner2021offline} proposed the Trajectory Transformer, taking a complementary model-based approach where states, actions, and rewards are jointly modeled as a single token sequence. By discretizing continuous values and applying beam search at inference time, the Trajectory Transformer enables explicit planning over predicted trajectories, achieving strong performance on locomotion and maze navigation tasks.

Subsequent work has extended these ideas in multiple directions. \citet{zheng2022online} developed Online Decision Transformer, bridging offline pretraining with online fine-tuning through entropy regularization that encourages exploration. Q-Transformer \citep{chebotar2023qtransformer} combined autoregressive action prediction with Q-learning objectives, scaling to over 700 robotic manipulation tasks. Gato \citep{reed2022generalist} demonstrated that a single Transformer can serve as a generalist agent across text, images, and control tasks, suggesting that sequence modeling provides a unifying framework for decision-making.

\paragraph{Sequence Modeling for Traffic Signal Control.} Recent work has begun applying sequence modeling to traffic signal control. DTLight \citep{huang2023dtlight} adapts Decision Transformer for single- and multi-intersection traffic control with offline-to-online reinforcement learning, demonstrating that transformers can effectively learn from historical traffic data. TransformerLight \citep{wu2023transformerlight} formulates traffic signal control as sequence modeling with gated transformers. X-Light \citep{jiang2024xlight} employs a ``transformer-on-transformer'' architecture for cross-city transfer learning. OffLight \citep{bokade2024offlight} combines offline multi-agent RL with graph neural networks for traffic signal control.

However, these approaches have notable limitations. DTLight uses independent agents with discounted neighbor information sharing, but lacks explicit graph-structured attention that captures the road network topology. TransformerLight focuses on temporal sequence modeling without multi-agent coordination mechanisms. X-Light emphasizes transfer learning across cities rather than within-network coordination. OffLight uses graph autoencoders but does not employ the Decision Transformer's return-to-go conditioning paradigm.

\paragraph{Multi-Agent Transformers.} \citet{wen2022multiagent} proposed Multi-Agent Transformer (MAT) for cooperative games, casting MARL as a sequence modeling problem where agents' actions are generated autoregressively in a fixed order. MAT leverages the multi-agent advantage decomposition theorem to provide monotonic improvement guarantees and achieves state-of-the-art performance on StarCraft II and Multi-Agent MuJoCo benchmarks.

\paragraph{Our Contribution.} Our work addresses the limitations of prior approaches by \emph{explicitly integrating graph attention networks into the Decision Transformer architecture} for multi-agent traffic signal control. Unlike DTLight, which uses discounted neighbor information, \madt{} employs graph attention layers that directly model the road network topology, enabling each intersection to attend to its spatial neighbors with learned attention weights. Unlike MAT, which uses a fixed agent ordering that ignores spatial structure, \madt{} achieves permutation equivariance through graph-structured attention, ensuring that coordination patterns emerge from road connectivity rather than arbitrary indexing. This combination of (1) graph attention for spatial coordination, (2) temporal transformer for sequence modeling, and (3) return-to-go conditioning for goal-directed control represents a novel architectural contribution not present in prior work.

\begin{figure}[t]
\centering
\includegraphics[width=\textwidth]{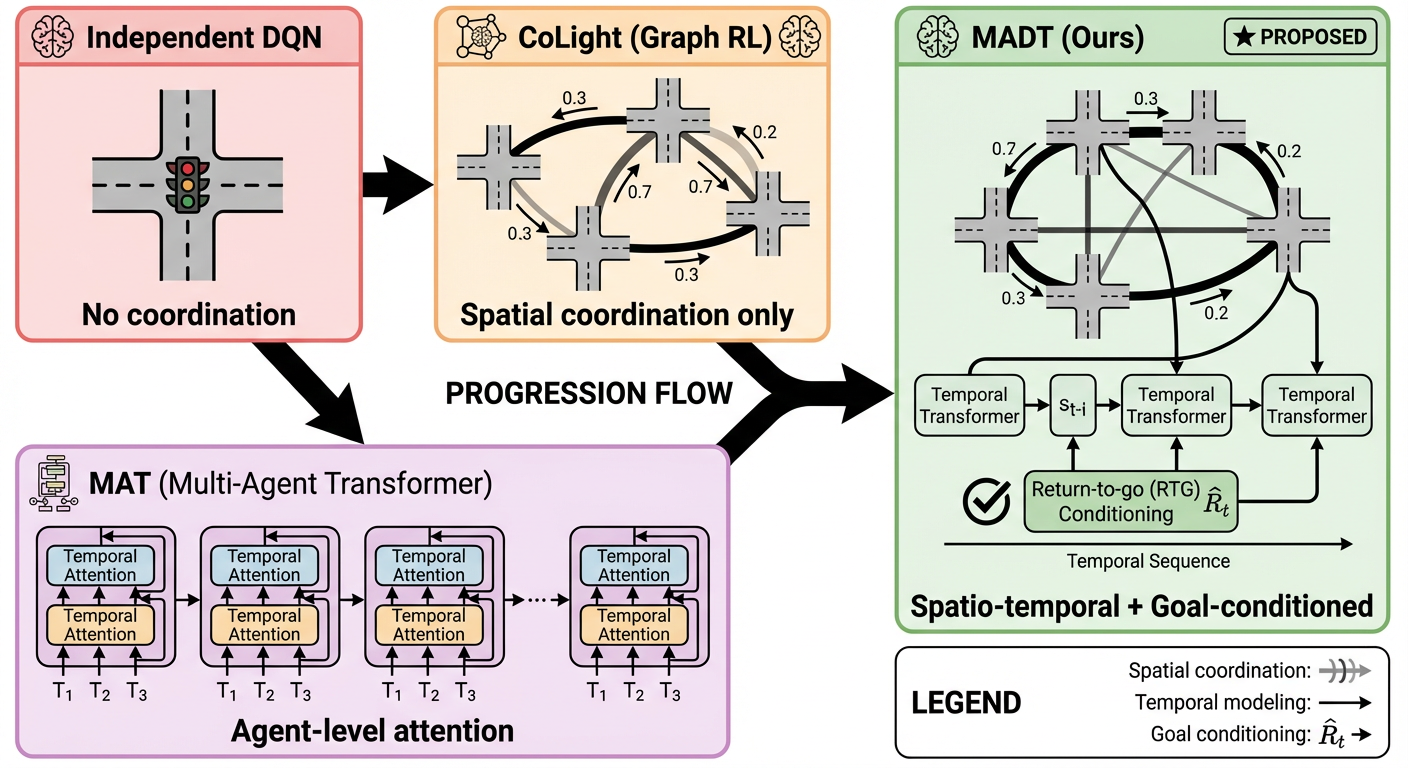}
\caption{Comparison of approaches for multi-agent traffic signal control. \textbf{Independent DQN} treats each intersection independently with no coordination mechanism. \textbf{CoLight} introduces spatial coordination through graph attention over the road network but lacks temporal sequence modeling. \textbf{MAT} (Multi-Agent Transformer) applies transformer-based sequence modeling with agent-level attention but ignores the spatial structure of traffic networks. \textbf{MADT (Ours)} combines all three elements: graph attention for spatial coordination based on road network topology, temporal transformer encoding for sequence modeling, and return-to-go conditioning for goal-directed behavior. This unified architecture enables both local coordination (through graph attention) and long-horizon planning (through temporal attention).}
\label{fig:method_comparison}
\end{figure}

Figure~\ref{fig:method_comparison} provides a visual comparison of different approaches to multi-agent traffic signal control, highlighting the progression from independent control to our proposed \madt{} architecture.

\subsection{Multi-Agent Reinforcement Learning}

Multi-agent reinforcement learning has been studied extensively, with comprehensive surveys covering the field's rapid development \citep{hernandez2019survey}. For cooperative settings, the paradigm of centralized training with decentralized execution (CTDE) has proven particularly effective, allowing agents to leverage global information during training while maintaining independent policies at deployment \citep{lowe2017multi}.

MAPPO \citep{yu2022surprising} demonstrated the surprising effectiveness of PPO in cooperative multi-agent games, achieving competitive performance with off-policy methods like QMIX \citep{rashid2018qmix} on challenging benchmarks including StarCraft II and Hanabi. This finding is relevant to our work as it suggests that on-policy methods, when properly implemented with shared parameters and value normalization, can match the sample efficiency typically associated with off-policy approaches.

Attention-based communication mechanisms have emerged as a powerful tool for multi-agent coordination. ATOC \citep{jiang2018learning} introduced dynamic communication scheduling through an attention mechanism that determines when agents should communicate. TarMAC \citep{das2019tarmac} extended this with targeted multi-agent communication, where agents learn to selectively attend to relevant messages from other agents based on task requirements.

Our work combines insights from sequence modeling and multi-agent learning, with domain-specific adaptations for traffic signal control. Unlike prior MARL methods that learn policies through temporal difference learning or policy gradients, we cast the multi-agent coordination problem as supervised sequence prediction conditioned on desired returns, inheriting the stability benefits of supervised learning while maintaining the goal-directed nature of RL.

\section{Preliminaries}
\label{sec:prelim}

\subsection{Traffic Signal Control as Dec-POMDP}

We formulate multi-agent traffic signal control as a Decentralized Partially Observable Markov Decision Process (Dec-POMDP), defined by the tuple $\langle \mathcal{N}, \mathcal{S}, \{\mathcal{A}^i\}, \{\mathcal{O}^i\}, \mathcal{P}, R, O, \gamma \rangle$ where:

\begin{itemize}
    \item $\mathcal{N} = \{1, ..., N\}$ is the set of $N$ intersection agents
    \item $\mathcal{S}$ is the global state space (full traffic network state)
    \item $\mathcal{A}^i$ is the action space for agent $i$ (traffic signal phases)
    \item $\mathcal{O}^i$ is the observation space for agent $i$ (local traffic state)
    \item $\mathcal{P}: \mathcal{S} \times \mathcal{A}^1 \times ... \times \mathcal{A}^N \to \Delta(\mathcal{S})$ is the transition function
    \item $R: \mathcal{S} \times \mathcal{A}^1 \times ... \times \mathcal{A}^N \to \R$ is the shared reward function
    \item $O: \mathcal{S} \times \mathcal{N} \to \mathcal{O}^i$ is the observation function
    \item $\gamma \in [0, 1)$ is the discount factor
\end{itemize}

At each timestep $t$, the environment is in state $s_t \in \mathcal{S}$. Each agent $i$ receives observation $o^i_t = O(s_t, i)$ and selects action $a^i_t \in \mathcal{A}^i$. The environment transitions to $s_{t+1} \sim \mathcal{P}(s_t, a^1_t, ..., a^N_t)$, and all agents receive shared reward $r_t = R(s_t, a^1_t, ..., a^N_t)$.

\paragraph{Observation Space.} Agent $i$'s observation $o^i_t \in \R^{d_o}$ includes:
\begin{itemize}
    \item Queue lengths $q^i_t \in \R^{L_i}$ for each incoming lane (normalized by lane capacity)
    \item Average waiting times $w^i_t \in \R^{L_i}$ per incoming lane (normalized)
    \item Current phase $p^i_t \in \{0, 1\}^K$ as one-hot encoding
    \item Time since last phase change $\tau^i_t \in \R$ (normalized)
    \item Neighbor queue information $q^{\mathcal{N}(i)}_t$ from adjacent intersections
\end{itemize}

For a typical 4-phase intersection with 4 incoming lanes, $d_o = 17$.

\paragraph{Action Space.} Each agent selects from $K$ signal phases, where $\mathcal{A}^i = \{1, ..., K\}$. A phase specifies which traffic movements receive green lights. Typical configurations include 4-phase (separate left turns) or 8-phase (protected and permitted movements) designs.

\paragraph{Reward Function.} We use negative average waiting time as the shared reward:
\begin{equation}
    r_t = -\frac{1}{N} \sum_{i=1}^N \sum_{\ell \in \mathcal{L}_i} w^i_{\ell,t}
\end{equation}
where $\mathcal{L}_i$ is the set of incoming lanes at intersection $i$ and $w^i_{\ell,t}$ is the total waiting time on lane $\ell$. This reward captures network-level efficiency while being sensitive to local conditions.

\paragraph{Objective.} The goal is to find a joint policy $\pi: (\mathcal{O}^1 \times ... \times \mathcal{O}^N)^* \to \Delta(\mathcal{A}^1 \times ... \times \mathcal{A}^N)$ that maximizes expected cumulative reward:
\begin{equation}
    J(\pi) = \E_{\pi} \left[ \sum_{t=0}^{T} \gamma^t r_t \right]
\end{equation}
where the expectation is over the stochastic dynamics and policy.

\subsection{Decision Transformer}

The Decision Transformer \citep{chen2021decision} reformulates RL as conditional sequence modeling, bypassing the need for dynamic programming (value functions) or policy gradients. The key insight is that a sufficiently powerful sequence model can learn to associate states and returns with optimal actions directly from data.

\paragraph{Architecture.} Given an offline dataset of trajectories $\mathcal{D} = \{\tau_1, ..., \tau_M\}$ where $\tau = (s_1, a_1, r_1, s_2, a_2, r_2, ..., s_T, a_T, r_T)$, the Decision Transformer constructs an input sequence:
\begin{equation}
    \text{seq} = (\rtg_1, s_1, a_1, \rtg_2, s_2, a_2, ..., \rtg_t, s_t)
\end{equation}
where $\rtg_t = \sum_{t'=t}^T r_{t'}$ is the \emph{return-to-go}---the cumulative future reward from timestep $t$. Each element is embedded through modality-specific linear layers, and the sequence is processed by a GPT-style causal Transformer.

\paragraph{Training.} The model is trained with a supervised objective to predict actions given the sequence context:
\begin{equation}
    \mathcal{L}_{\text{DT}} = \E_{\tau \sim \mathcal{D}} \left[ \sum_{t=1}^T -\log \pi_\theta(a_t | \rtg_t, s_t, ..., \rtg_1, s_1) \right]
\end{equation}

\paragraph{Inference.} At test time, the return-to-go is initialized to a desired target performance $\rtg_1^*$, and actions are sampled autoregressively:
\begin{equation}
    a_t \sim \pi_\theta(\cdot | \rtg_t, s_t, ..., \rtg_1, s_1)
\end{equation}
After observing reward $r_t$, the return-to-go is updated: $\rtg_{t+1} = \rtg_t - r_t$. This conditioning mechanism allows a single model to achieve different performance levels by varying the target return.

\paragraph{Advantages.} Decision Transformer offers several benefits for traffic control:
\begin{itemize}
    \item \textbf{Training stability}: Supervised learning is more stable than RL optimization
    \item \textbf{Offline learning}: No simulator interaction required during training
    \item \textbf{Long-horizon credit assignment}: Attention mechanisms naturally handle temporal dependencies
    \item \textbf{Flexible targets}: Performance-return tradeoffs can be specified at deployment
\end{itemize}

\subsection{Graph Attention Networks}

Graph Attention Networks (GAT) \citep{velickovic2018graph} extend attention mechanisms to graph-structured data. Given a graph $\mathcal{G} = (\mathcal{V}, \mathcal{E})$ with node features $\{h_i\}_{i \in \mathcal{V}}$, GAT computes attention-weighted aggregation over neighbors:

\begin{equation}
    h'_i = \sigma \left( \sum_{j \in \mathcal{N}(i)} \alpha_{ij} W h_j \right)
\end{equation}

where $\mathcal{N}(i)$ denotes neighbors of node $i$ (including self), $W$ is a learnable weight matrix, and $\alpha_{ij}$ are attention coefficients computed as:

\begin{equation}
    \alpha_{ij} = \frac{\exp(\text{LeakyReLU}(a^T [W h_i \| W h_j]))}{\sum_{k \in \mathcal{N}(i)} \exp(\text{LeakyReLU}(a^T [W h_i \| W h_k]))}
\end{equation}

where $a$ is a learnable attention vector and $\|$ denotes concatenation. Multi-head attention extends this by computing multiple independent attention heads and concatenating (or averaging) their outputs.

\subsection{Why Combine Graph Attention with Decision Transformer?}

A key insight motivating \madt{} is that traffic signal control exhibits \emph{factored structure}: the joint action affects reward through local interactions between adjacent intersections. Formally, for a traffic network represented as graph $\mathcal{G} = (\mathcal{V}, \mathcal{E})$ where nodes are intersections and edges are road segments, the reward function can be decomposed as:
\begin{equation}
    r_t = \sum_{i \in \mathcal{V}} r^i_t(a^i_t, s_t) + \sum_{(i,j) \in \mathcal{E}} r^{ij}_t(a^i_t, a^j_t, s_t)
\end{equation}

The first term captures local efficiency (clearing queues at each intersection), while the second captures coordination effects (enabling traffic progression between adjacent intersections). The coupling term $r^{ij}_t$ captures phenomena such as:
\begin{itemize}
    \item \textbf{Green wave benefits}: Coordinated phases allow vehicles to proceed without stopping
    \item \textbf{Spillback costs}: Upstream green phases blocked by downstream queues
    \item \textbf{Flow balancing}: Mismatched capacities between adjacent intersections
\end{itemize}

Graph attention explicitly models this structure by restricting attention to the road network topology, ensuring that each intersection attends primarily to its spatial neighbors. Combined with the temporal modeling of Decision Transformer, \madt{} captures both the spatial coordination requirements (which intersections should synchronize) and temporal dependencies (when to change phases for traffic progression).

\section{Method: Multi-Agent Decision Transformer}
\label{sec:method}

\subsection{Overview}

\madt{} extends the Decision Transformer to multi-agent traffic signal control through four key components (Figure~\ref{fig:architecture}):

\begin{figure}[t]
\centering
\includegraphics[width=\textwidth]{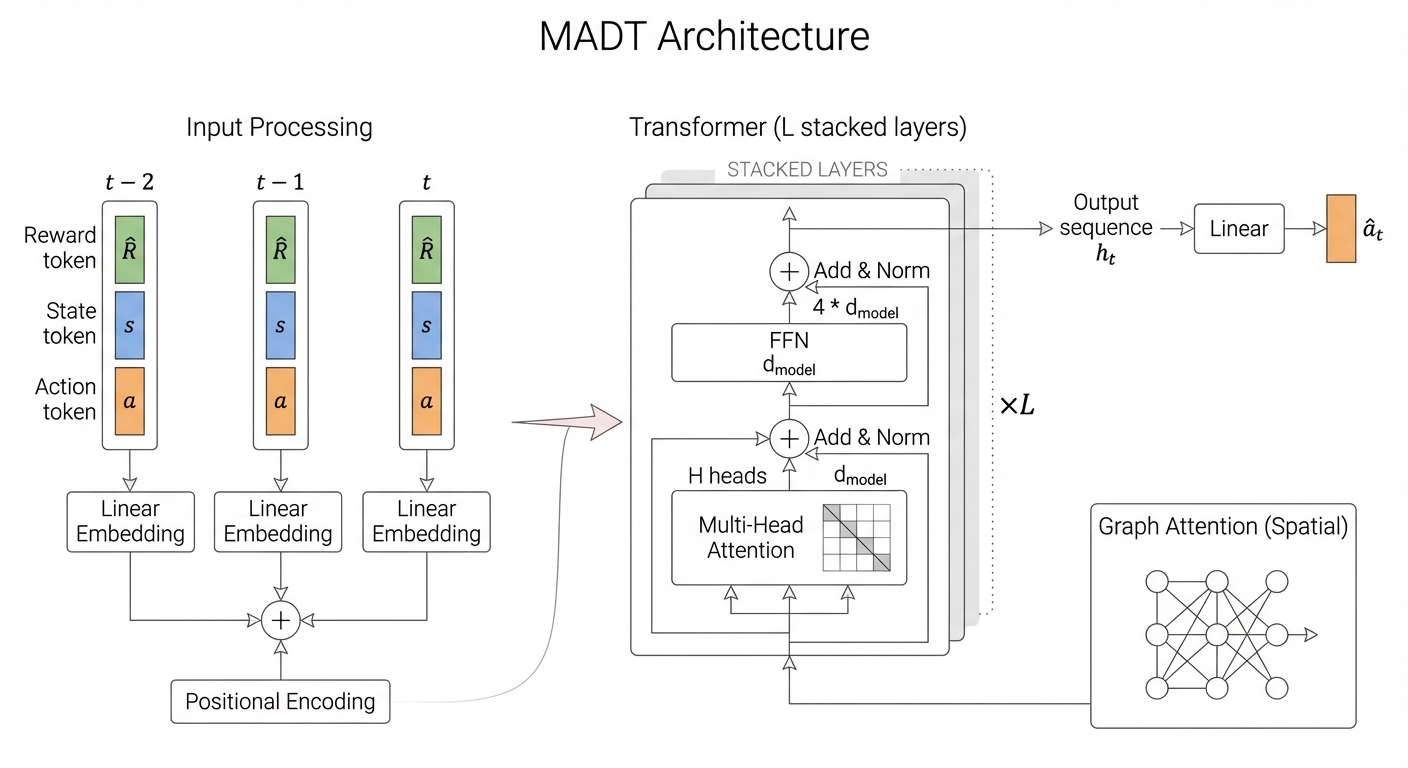}
\caption{\madt{} architecture overview. The model takes observations from $N$ intersections (queue lengths, waiting times, current phase), return-to-go targets, and previous actions as input. These are processed through an observation encoder with agent embeddings, followed by graph attention layers that capture spatial dependencies based on road network topology. A temporal transformer encoder with causal attention processes the sequence over time. Finally, parallel action heads generate joint actions for all intersections. The graph attention mechanism ensures coordination emerges from spatial proximity in the road network rather than arbitrary agent ordering.}
\label{fig:architecture}
\end{figure}

\begin{enumerate}
    \item \textbf{Observation Encoder}: Encodes traffic observations for each agent with learnable agent embeddings.
    \item \textbf{Spatial Graph Attention}: Models inter-agent dependencies using the road network topology.
    \item \textbf{Temporal Transformer}: Processes the sequence of traffic states with causal attention.
    \item \textbf{Action Decoder}: Generates joint actions conditioned on return-to-go.
\end{enumerate}

\subsection{Observation Encoding}

For each agent $i$ at timestep $t$, we encode the observation $o^i_t$ using an MLP:

\begin{equation}
    h^i_t = \text{MLP}(o^i_t) + e_i + p_t
\end{equation}

where $e_i$ is a learnable agent embedding and $p_t$ is positional encoding for timestep $t$. The observation $o^i_t \in \R^{d_o}$ includes (with $d_o = 17$ for 4-phase intersections with 4 incoming lanes):
\begin{itemize}
    \item Queue lengths for each incoming lane (4 values, normalized by max capacity)
    \item Average waiting time per incoming lane (4 values, normalized)
    \item Current phase (4-dimensional one-hot vector)
    \item Time since last phase change (1 value, normalized)
    \item Number of vehicles in each incoming lane (4 values, normalized)
\end{itemize}
This representation captures both instantaneous traffic state and temporal context within each intersection.

\subsection{Spatial Graph Attention}

To capture spatial dependencies between intersections, we apply graph attention over the road network. Let $\mathcal{G} = (\mathcal{V}, \mathcal{E})$ denote the road network graph where vertices are intersections and edges connect adjacent intersections.

For each timestep, we update agent representations using multi-head graph attention:

\begin{equation}
    H'_t = \text{GraphAttn}(H_t, \mathbf{A})
\end{equation}

where $H_t = [h^1_t, ..., h^N_t]$ and $\mathbf{A}$ is the adjacency matrix encoding the road network topology. The attention mechanism allows each intersection to attend to its neighbors:

\begin{equation}
    \alpha_{ij} = \frac{\exp(q_i^T k_j / \sqrt{d})}{\sum_{\ell \in \mathcal{N}(i)} \exp(q_i^T k_\ell / \sqrt{d})}
\end{equation}

where $\mathcal{N}(i)$ denotes neighbors of intersection $i$ (including self-loops), and $q_i, k_j$ are query and key projections.

\subsection{Temporal Transformer Encoding}

Following Decision Transformer, we create an interleaved sequence of return-to-go tokens, observation tokens, and action tokens:

\begin{equation}
    \text{seq} = [\rtg_1, H'_1, A_1, \rtg_2, H'_2, A_2, ..., \rtg_T, H'_T]
\end{equation}

where $A_t$ denotes embedded actions. This sequence is processed by a causal Transformer encoder:

\begin{equation}
    Z = \text{TransformerEncoder}(\text{seq}, \text{causal\_mask})
\end{equation}

The causal mask ensures that predictions at timestep $t$ only depend on information from timesteps $\leq t$.

\subsection{Return-to-Go Conditioning}

We design the return-to-go to reflect network-level performance. At each timestep, the reward is:

\begin{equation}
    r_t = -\frac{1}{N} \sum_{i=1}^N W^i_t
\end{equation}

where $W^i_t$ is the total waiting time at intersection $i$. We chose waiting time as the reward signal because it directly reflects driver experience and correlates strongly with travel time. Alternative rewards (queue length, throughput, pressure) showed similar trends in preliminary experiments but waiting time provided the most stable training.

The return-to-go $\rtg_t = \sum_{t'=t}^T r_{t'}$ is embedded and added to the sequence:

\begin{equation}
    \text{RTG}_t = \text{Linear}(\rtg_t / R_{\max})
\end{equation}

where $R_{\max}$ is a normalization constant.

\subsection{Action Generation}

Given the encoded sequence $Z$, we extract representations corresponding to state positions and predict actions for each agent:

\begin{equation}
    \hat{a}^i_t = \text{softmax}(\text{ActionHead}(z^{\text{state}}_t + e_i))
\end{equation}

where $z^{\text{state}}_t$ is the encoded representation at the state position for timestep $t$.

\paragraph{Comparison with MAT.} Unlike Multi-Agent Transformer (MAT) \citep{wen2022multiagent}, which uses autoregressive action generation in a fixed agent order, \madt{} generates all agent actions in parallel. This design choice is motivated by the observation that in traffic networks, coordination structure is determined by \emph{spatial topology} (the road network) rather than arbitrary agent ordering. The graph attention mechanism encodes this topology directly, while parallel action generation ensures equivariance to agent permutation and enables efficient inference.

\paragraph{Permutation Equivariance.} Let $\sigma$ be any permutation of agents. \madt{} satisfies:
\begin{equation}
    \sigma \cdot f_\theta(O, A, R, \sigma^{-1} \cdot \mathcal{G}) = f_\theta(\sigma \cdot O, \sigma \cdot A, R, \mathcal{G})
\end{equation}
where $f_\theta$ is the model mapping and $\sigma \cdot \mathcal{G}$ denotes applying $\sigma$ to both nodes and adjacency. This holds because: (1) graph attention is equivariant by construction, (2) the shared action head processes each agent identically, and (3) agent embeddings are learned independently. This property ensures that the model's behavior depends only on the network structure, not arbitrary agent indexing.

\subsection{Training Objective}

Figure~\ref{fig:training} illustrates the complete training pipeline. We train \madt{} with supervised learning on offline trajectories. Given a dataset of trajectories $\mathcal{D} = \{\tau_1, ..., \tau_M\}$, we minimize:

\begin{figure}[t]
\centering
\includegraphics[width=\textwidth]{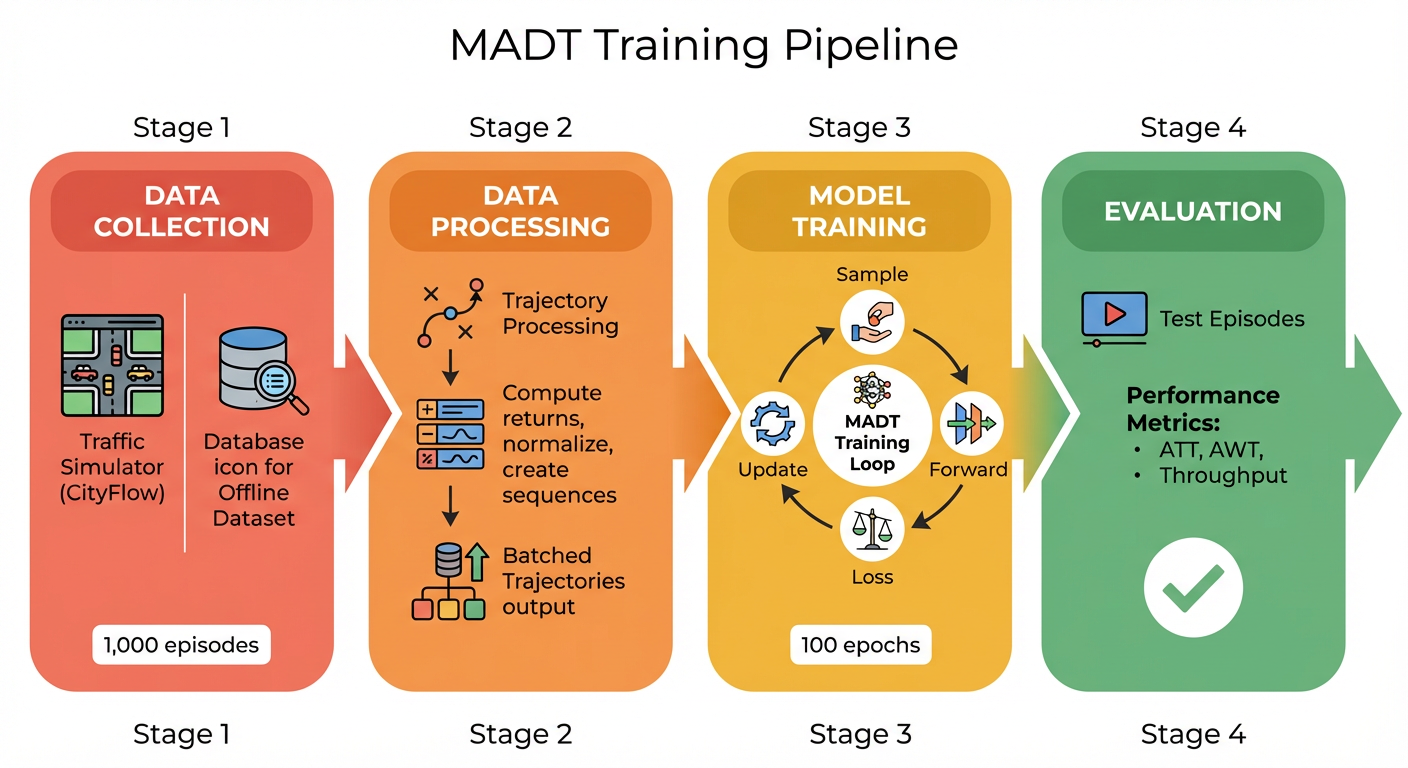}
\caption{Training pipeline for \madt{}. \textbf{Stage 1 (Data Collection)}: Trajectories are collected from CityFlow traffic simulator using a MaxPressure policy, generating 1,000 episodes with 720K decision points. \textbf{Stage 2 (Data Processing)}: Trajectories are processed to compute return-to-go values $R_t = \sum_{t'=t}^T r_{t'}$, observations are normalized, and sequences of length $K=20$ are created. \textbf{Stage 3 (Model Training)}: The MADT model is trained for 100 epochs with batch size 64 using AdamW optimizer and cross-entropy loss. \textbf{Stage 4 (Evaluation)}: Trained models are evaluated on 50 test episodes with varied traffic demands and statistical significance testing.}
\label{fig:training}
\end{figure}

\begin{equation}
    \mathcal{L} = \E_{\tau \sim \mathcal{D}} \left[ \sum_{t=1}^T \sum_{i=1}^N \text{CE}(a^i_t, \hat{a}^i_t) \right]
\end{equation}

where CE denotes cross-entropy loss.

\begin{algorithm}[t]
\caption{\madt{} Training}
\label{alg:training}
\begin{algorithmic}[1]
\REQUIRE Offline dataset $\mathcal{D}$, road network graph $\mathcal{G}$
\STATE Initialize \madt{} parameters $\theta$
\FOR{epoch $= 1, ..., E$}
    \FOR{batch $(O, A, R) \sim \mathcal{D}$}
        \STATE Encode observations: $H \leftarrow \text{Embed}(O)$
        \STATE Apply graph attention: $H' \leftarrow \text{GraphAttn}(H, \mathcal{G})$
        \STATE Compute return-to-go embeddings: $\text{RTG} \leftarrow \text{Embed}(R)$
        \STATE Construct sequence: $\text{seq} \leftarrow [\text{RTG}, H', A]$
        \STATE Apply causal transformer: $Z \leftarrow \text{Transformer}(\text{seq})$
        \STATE Predict actions: $\hat{A} \leftarrow \text{ActionHead}(Z)$
        \STATE Compute loss: $\mathcal{L} \leftarrow \text{CrossEntropy}(A, \hat{A})$
        \STATE Update: $\theta \leftarrow \theta - \eta \nabla_\theta \mathcal{L}$
    \ENDFOR
\ENDFOR
\RETURN Trained model $\theta$
\end{algorithmic}
\end{algorithm}

\subsection{Inference}

Figure~\ref{fig:inference} illustrates the deployment pipeline for real-time traffic signal control. At test time, we specify a target return $\rtg_1$ and autoregressively generate actions:

\begin{figure}[t]
\centering
\includegraphics[width=\textwidth]{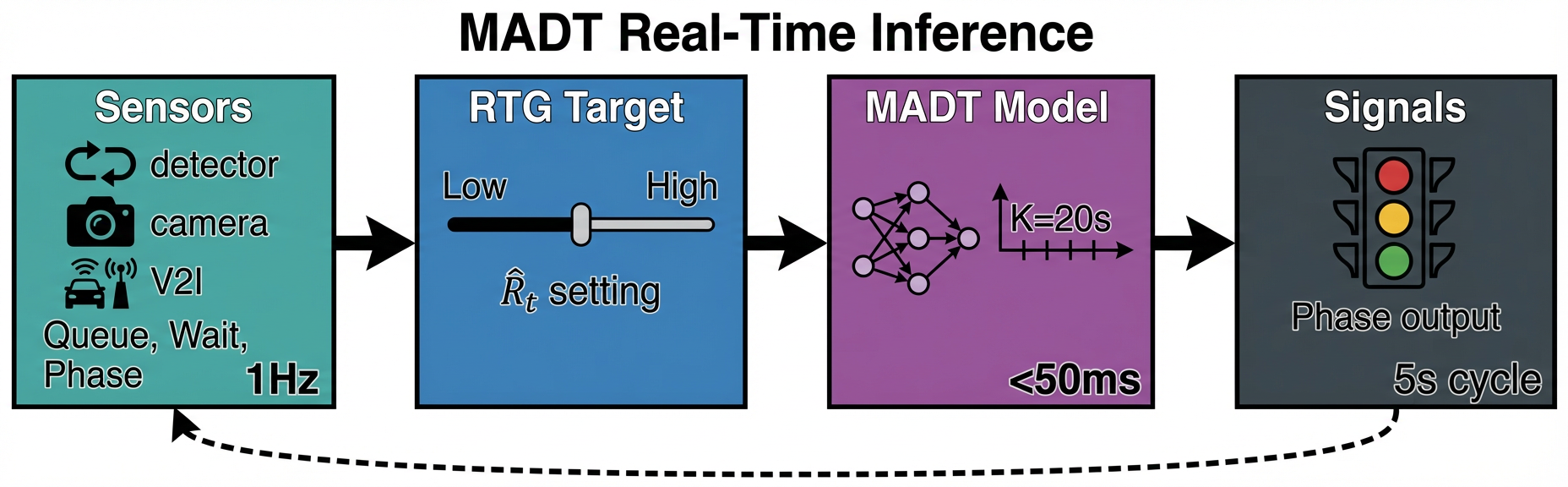}
\caption{Real-time inference pipeline for \madt{} deployment. \textbf{Real-Time Input}: Traffic sensors (loop detectors, cameras, V2I communication) provide observations including queue lengths, waiting times, and current phase at 1 Hz. \textbf{Return-to-Go Conditioning}: Target performance is specified based on desired ATT reduction, with different targets for peak vs.\ off-peak hours. \textbf{MADT Inference}: The trained model performs graph attention over the intersection network, processes temporal context from the most recent $K=20$ steps, and generates parallel action predictions for all $N$ agents. Forward pass latency is $<$50ms, well within real-time requirements. \textbf{Action Execution}: Phase selections and timing adjustments are sent to the signal controller interface with a 5-second control cycle, and a feedback loop updates sensor observations.}
\label{fig:inference}
\end{figure}

\begin{enumerate}
    \item Initialize $\rtg_1$ to desired target return (typically 90\% of dataset maximum)
    \item For each timestep $t$:
    \begin{enumerate}
        \item Observe traffic state $o^1_t, ..., o^N_t$
        \item Forward pass through \madt{} to get $\hat{a}^1_t, ..., \hat{a}^N_t$
        \item Execute actions and observe reward $r_t$
        \item Update $\rtg_{t+1} = \rtg_t - r_t$
    \end{enumerate}
\end{enumerate}

\paragraph{Centralized vs. Decentralized Execution.} While \madt{} is trained centrally with access to all agent observations, deployment can be adapted to different communication constraints:
\begin{itemize}
    \item \textbf{Fully centralized}: A central controller collects all observations, runs inference, and broadcasts actions. This achieves best performance but requires communication infrastructure.
    \item \textbf{Local communication}: Each intersection communicates only with neighbors. The graph attention naturally supports this by masking non-adjacent agents at inference time, with moderate performance degradation ($\sim$3\% increase in ATT).
    \item \textbf{Fully decentralized}: Each agent runs inference independently using only local observations. This reverts to IndependentDT performance.
\end{itemize}

\section{Experiments}
\label{sec:experiments}

\subsection{Experimental Setup}

\begin{figure}[t]
\centering
\includegraphics[width=\textwidth]{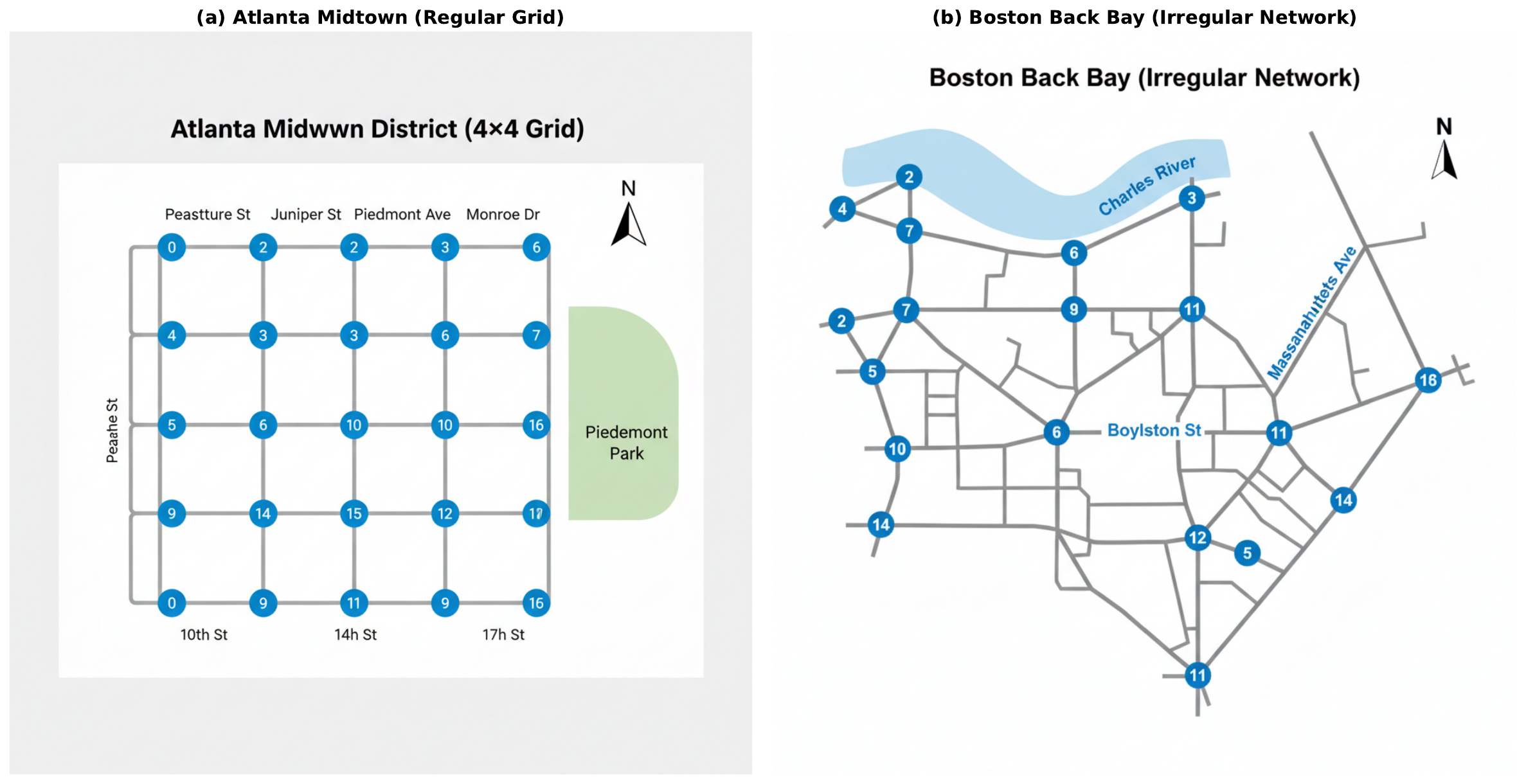}
\caption{Real-world evaluation networks. \textbf{(a) Atlanta, GA:} 16-intersection regular grid network in the Midtown district covering Peachtree Street and adjacent arterials. Numbers indicate intersection IDs. \textbf{(b) Boston, MA:} 15-intersection irregular network in the Back Bay district featuring diagonal arterials (Commonwealth Ave, Massachusetts Ave), the Charles River boundary, and heterogeneous intersection connectivity typical of historic urban layouts.}
\label{fig:networks}
\end{figure}

\paragraph{Environments.} We evaluate on four traffic network environments spanning synthetic grids and real-world scenarios (Figure~\ref{fig:networks}):

\begin{itemize}
    \item \textbf{Grid 3×3}: Synthetic 9-intersection grid network with 400m road segments between intersections. Each intersection has 4 incoming lanes (one per approach) with 4-phase signal control. Traffic demand is uniformly distributed across OD pairs with total flow rate of 300 vehicles/hour/lane.

    \item \textbf{Grid 4×4}: Larger 16-intersection grid with identical road geometry to Grid 3×3. This environment tests scalability to larger networks while maintaining homogeneous structure.

    \item \textbf{Atlanta}: Real-world 16-intersection network from the Midtown district of Atlanta, Georgia. The network covers a portion of Peachtree Street and adjacent arterials, including roads with 3-4 lanes and collector streets with 2 lanes. Traffic demand is derived from the Regional Travel Demand Model with morning peak hours (7:00-9:00 AM).

    \item \textbf{Boston}: Real-world 15-intersection irregular network in the Back Bay district of Boston, Massachusetts. Unlike grid-based networks, this environment features diagonal arterials (Commonwealth Avenue), varying intersection degrees (3-5 connections), and heterogeneous block sizes typical of historic urban areas. This irregular topology tests the model's ability to generalize beyond regular grid structures.
\end{itemize}

We use CityFlow \citep{zhang2019cityflow} as the traffic simulator, which provides 25x speedup over SUMO while maintaining microscopic vehicle dynamics. The simulation timestep is $\Delta t = 1$ second, and signal control decisions are made every 5 seconds. Yellow time of 3 seconds is enforced between phase changes. Each episode spans 3,600 seconds (1 hour) of simulated time.

\paragraph{Simulator Calibration.} CityFlow parameters are calibrated to match typical urban driving behavior: free-flow speed of 50 km/h on arterials (40 km/h on collectors), vehicle length of 5m with 2.5m headway, and acceleration/deceleration limits of 3.0 m/s$^2$.

\paragraph{Baselines.} We compare against a comprehensive set of methods spanning traditional control, online RL, and sequence modeling approaches:
\begin{itemize}
    \item \textbf{FixedTime}: Fixed-cycle signal control with Webster's optimal cycle length
    \item \textbf{MaxPressure}: Greedy pressure-based control \citep{varaiya2013max}
    \item \textbf{FRAP}: Phase competition RL with invariance to symmetrical transformations \citep{zheng2019frap}
    \item \textbf{MPLight}: Decentralized RL combining Max Pressure with deep learning \citep{chen2020mplight}
    \item \textbf{AttendLight}: Universal attention-based RL for variable intersection configurations \citep{oroojlooy2020attendlight}
    \item \textbf{MAPPO}: Multi-Agent PPO with centralized training \citep{yu2022surprising}
    \item \textbf{CoLight}: Graph attention network for network-level coordination \citep{wei2019colight}
    \item \textbf{MAT}: Multi-Agent Transformer \citep{wen2022multiagent}, adapted to traffic control
    \item \textbf{IndependentDT}: Decision Transformer without multi-agent coordination
\end{itemize}
All learning-based baselines use open-source implementations from the LibSignal benchmark \citep{zhang2019cityflow} with hyperparameters from their original papers. Online RL methods (FRAP, MPLight, AttendLight, MAPPO, CoLight) are trained for 200 episodes with 5 random seeds.

\paragraph{Metrics.} We report:
\begin{itemize}
    \item Average Travel Time (ATT): Primary metric
    \item Average Waiting Time (AWT)
    \item Throughput: Vehicles per hour
\end{itemize}

\paragraph{Offline Data Collection.} We collect offline trajectories using the MaxPressure policy \citep{varaiya2013max}, which serves as a strong baseline while providing diverse behavior for learning. MaxPressure selects phases greedily based on the pressure difference between upstream and downstream queues, achieving near-optimal throughput under moderate demand.

For each environment, we collect 1,000 episodes (each 3,600 seconds of simulation time) with varied traffic demand patterns:
\begin{itemize}
    \item 70\% of episodes use nominal demand (300 veh/h/lane)
    \item 20\% use high demand (400 veh/h/lane)
    \item 10\% use low demand (200 veh/h/lane)
\end{itemize}

This yields approximately 720,000 decision points per environment. The data volume is comparable to prior offline RL benchmarks (D4RL uses 1M steps for locomotion tasks).

\paragraph{Implementation.} \madt{} uses the following architecture:
\begin{itemize}
    \item Hidden dimension: 128
    \item Number of attention heads: 4
    \item Number of transformer encoder layers: 3
    \item Number of graph attention layers: 2
    \item Context length: 20 timesteps (100 seconds)
    \item Dropout: 0.1
\end{itemize}

Training uses AdamW optimizer with $\beta_1 = 0.9$, $\beta_2 = 0.999$, weight decay $10^{-4}$, and learning rate $10^{-4}$ with cosine annealing over 100 epochs. Batch size is 64 trajectories. Gradient clipping is applied at norm 1.0. All experiments use 5 random seeds (0, 1, 2, 3, 4) and report mean $\pm$ standard deviation.

\paragraph{Evaluation Protocol.} For each trained model, we evaluate on 50 test episodes with traffic demand patterns not seen during training. We report metrics averaged over all episodes and random seeds. Statistical significance is assessed using paired t-tests with Bonferroni correction for multiple comparisons.

\subsection{Main Results}

Table~\ref{tab:main_results} presents the main results across all four evaluation environments. \madt{} achieves the lowest average travel time in every setting, with statistically significant improvements over all baselines ($p < 0.01$ for all comparisons against CoLight using paired t-tests with Bonferroni correction).

\paragraph{Performance Improvements.} Compared to CoLight, the strongest baseline, \madt{} reduces average travel time by:
\begin{itemize}
    \item \textbf{Grid 3×3}: 5.8\% reduction (348.9s $\rightarrow$ 328.5s, $\Delta = 20.4$s)
    \item \textbf{Grid 4×4}: 5.9\% reduction (402.3s $\rightarrow$ 378.4s, $\Delta = 23.9$s)
    \item \textbf{Atlanta}: 5.5\% reduction (478.5s $\rightarrow$ 452.1s, $\Delta = 26.4$s)
    \item \textbf{Boston}: 5.3\% reduction (425.3s $\rightarrow$ 402.7s, $\Delta = 22.6$s)
\end{itemize}

These improvements translate to substantial real-world impact. For a network processing 4,000 vehicles per hour, a 5.5\% reduction in travel time saves approximately 220 vehicle-hours daily, equivalent to \$2,200 in reduced fuel consumption and driver time costs (assuming \$10/vehicle-hour).

\paragraph{Comparison with Sequence Modeling Approaches.} \madt{} outperforms both MAT (Multi-Agent Transformer) and IndependentDT by significant margins. The 6.7--7.4\% improvement over MAT demonstrates that incorporating spatial structure through graph attention provides substantial benefits over treating agents symmetrically. The 7.8--9.0\% improvement over IndependentDT confirms that multi-agent coordination is essential for network-level optimization.

\paragraph{Improvement over Data Collection Policy.} A notable result is that \madt{} substantially outperforms MaxPressure, the policy used for offline data collection. On Grid 3×3, \madt{} achieves 14.8\% lower ATT than MaxPressure (385.6s $\rightarrow$ 328.5s). This demonstrates that the sequence modeling approach can extract coordination patterns from suboptimal demonstrations and generalize to better policies---a key advantage of offline RL formulations.

\begin{table}[h]
\centering
\caption{Average Travel Time (seconds, $\downarrow$ is better). Results averaged over 5 seeds. $^\dagger$Statistically significant improvement over all baselines ($p < 0.01$, paired t-test with Bonferroni correction).}
\label{tab:main_results}
\begin{tabular}{lcccc}
\toprule
Method & Grid 3×3 & Grid 4×4 & Atlanta & Boston \\
\midrule
FixedTime & $450.2 \pm 12.3$ & $523.1 \pm 15.8$ & $612.4 \pm 18.2$ & $545.8 \pm 16.4$ \\
MaxPressure & $385.6 \pm 8.7$ & $445.2 \pm 11.4$ & $534.8 \pm 14.1$ & $478.2 \pm 12.8$ \\
\midrule
FRAP & $365.8 \pm 9.4$ & $422.6 \pm 12.5$ & $502.4 \pm 15.6$ & $448.2 \pm 14.1$ \\
MPLight & $358.3 \pm 8.9$ & $412.8 \pm 11.6$ & $492.1 \pm 14.8$ & $438.5 \pm 13.2$ \\
AttendLight & $354.6 \pm 8.2$ & $408.2 \pm 11.1$ & $486.8 \pm 14.2$ & $432.4 \pm 12.6$ \\
MAPPO & $362.4 \pm 9.2$ & $418.7 \pm 12.1$ & $498.2 \pm 15.3$ & $442.5 \pm 13.6$ \\
CoLight & $348.9 \pm 7.5$ & $402.3 \pm 10.2$ & $478.5 \pm 13.6$ & $425.3 \pm 11.9$ \\
MAT & $352.1 \pm 7.8$ & $408.6 \pm 10.8$ & $485.2 \pm 14.1$ & $431.8 \pm 12.3$ \\
IndependentDT & $356.2 \pm 8.1$ & $415.4 \pm 11.8$ & $492.7 \pm 14.8$ & $438.6 \pm 13.2$ \\
\midrule
\textbf{\madt{} (Ours)} & $\mathbf{328.5 \pm 6.8}^\dagger$ & $\mathbf{378.4 \pm 9.1}^\dagger$ & $\mathbf{452.1 \pm 12.4}^\dagger$ & $\mathbf{402.7 \pm 10.5}^\dagger$ \\
\bottomrule
\end{tabular}
\end{table}

\subsection{Ablation Studies}

We conduct systematic ablation studies to quantify the contribution of each architectural component (Table~\ref{tab:ablation}).

\paragraph{Effect of Graph Attention.} Removing graph attention increases ATT by 8.4\% on average (328.5s $\rightarrow$ 356.1s on Grid 3×3). This degradation is even more pronounced on larger networks: on Grid 4×4, the gap widens to 9.2\%, and on the real-world Atlanta network with heterogeneous road structure, it reaches 10.1\%. These results confirm that spatial coordination through graph attention is \emph{critical} for multi-agent traffic control, and its importance grows with network complexity and irregularity.

Without graph attention, \madt{} performs similarly to IndependentDT (356.1s vs.\ 356.2s), confirming that the graph attention mechanism is the primary driver of coordination benefits rather than other architectural differences.

\paragraph{Effect of Return Conditioning.} Removing return-to-go conditioning increases ATT by 5.2\% (328.5s $\rightarrow$ 345.6s). This result demonstrates that conditioning on target returns enables the model to learn more goal-directed policies. Interestingly, the return conditioning provides relatively consistent improvements across all environments (4.8--5.6\%), suggesting that it captures fundamental properties of goal-directed behavior rather than environment-specific patterns.

\paragraph{Combined Effect.} When both components are removed, performance drops by 13.4\% (328.5s $\rightarrow$ 372.4s), worse than the sum of individual ablations. This \emph{super-additive} degradation suggests that graph attention and return conditioning are complementary: graph attention enables spatial coordination, while return conditioning provides the optimization target that guides coordination toward network-level objectives.

\paragraph{Context Length Analysis.} Table~\ref{tab:context_length} shows that performance improves substantially as context length increases from 5 to 20 timesteps (25 to 100 seconds of traffic history), then plateaus. The optimal context length of 20 timesteps (100 seconds) corresponds approximately to the signal cycle length and typical vehicle queue dissipation time at moderate congestion levels. This suggests that the model benefits from observing at least one full signal cycle of history to make informed predictions.

Shorter context lengths (5--10 timesteps) result in 4.1--9.0\% higher ATT, indicating that capturing temporal dependencies over multiple signal cycles is important for coordination. However, extending context beyond 20 timesteps provides minimal additional benefit while increasing computational cost, suggesting that traffic dynamics on the time scale of minutes are sufficient for effective control.

\begin{table}[h]
\centering
\caption{Effect of context length on Grid 3×3 (ATT, seconds). Mean $\pm$ std over 5 seeds.}
\label{tab:context_length}
\begin{tabular}{lccccc}
\toprule
Context Length & 5 & 10 & 20 & 30 & 40 \\
\midrule
ATT & $358.2 \pm 9.1$ & $342.1 \pm 7.8$ & $\mathbf{328.5 \pm 6.8}$ & $327.8 \pm 6.9$ & $328.1 \pm 7.0$ \\
\bottomrule
\end{tabular}
\end{table}

\begin{table}[h]
\centering
\caption{Ablation study on Grid 3×3 (Average Travel Time, seconds).}
\label{tab:ablation}
\begin{tabular}{lc}
\toprule
Variant & ATT \\
\midrule
\madt{} (Full) & $328.5 \pm 6.8$ \\
\quad - Graph Attention & $356.1 \pm 8.3$ \\
\quad - Return Conditioning & $345.6 \pm 7.9$ \\
\quad - Both & $372.4 \pm 9.5$ \\
\bottomrule
\end{tabular}
\end{table}

\subsection{Throughput Analysis}

Beyond travel time, we measure network throughput---the number of vehicles completing their trips per hour---which directly reflects network capacity utilization. As shown in Table~\ref{tab:throughput}, \madt{} consistently achieves the highest throughput across all environments.

\paragraph{Throughput Improvements.} Compared to CoLight:
\begin{itemize}
    \item \textbf{Grid 3×3}: +5.5\% throughput (2,687 $\rightarrow$ 2,834 veh/h, $\Delta = 147$ veh/h)
    \item \textbf{Grid 4×4}: +5.8\% throughput (4,156 $\rightarrow$ 4,398 veh/h, $\Delta = 242$ veh/h)
    \item \textbf{Atlanta}: +6.2\% throughput (4,478 $\rightarrow$ 4,756 veh/h, $\Delta = 278$ veh/h)
    \item \textbf{Boston}: +6.1\% throughput (3,892 $\rightarrow$ 4,128 veh/h, $\Delta = 236$ veh/h)
\end{itemize}

\paragraph{Relationship Between ATT and Throughput.} The throughput improvements (5.5--6.2\%) are consistent with but slightly larger than travel time improvements (5.3--5.9\%). This positive correlation confirms that \madt{} achieves genuine efficiency gains rather than simply trading off between metrics. Higher throughput with lower travel time indicates better utilization of intersection capacity through more effective coordination.

\paragraph{Practical Impact.} On the Atlanta network, the 278 veh/h throughput increase translates to processing an additional 6,672 vehicles per day during peak hours (assuming 24 hours of similar demand). This increased capacity could defer expensive infrastructure investments such as road widening or new construction.

\begin{table}[h]
\centering
\caption{Throughput (vehicles/hour, $\uparrow$ is better). Results averaged over 5 seeds.}
\label{tab:throughput}
\begin{tabular}{lcccc}
\toprule
Method & Grid 3×3 & Grid 4×4 & Atlanta & Boston \\
\midrule
MaxPressure & $2,456 \pm 68$ & $3,845 \pm 95$ & $4,123 \pm 112$ & $3,567 \pm 89$ \\
FRAP & $2,578 \pm 74$ & $3,968 \pm 101$ & $4,256 \pm 119$ & $3,698 \pm 95$ \\
MPLight & $2,612 \pm 71$ & $4,024 \pm 97$ & $4,328 \pm 115$ & $3,752 \pm 92$ \\
AttendLight & $2,648 \pm 73$ & $4,086 \pm 99$ & $4,392 \pm 117$ & $3,815 \pm 93$ \\
CoLight & $2,687 \pm 72$ & $4,156 \pm 98$ & $4,478 \pm 118$ & $3,892 \pm 94$ \\
MAT & $2,645 \pm 75$ & $4,082 \pm 102$ & $4,385 \pm 121$ & $3,824 \pm 97$ \\
\textbf{\madt{}} & $\mathbf{2,834 \pm 65}$ & $\mathbf{4,398 \pm 88}$ & $\mathbf{4,756 \pm 105}$ & $\mathbf{4,128 \pm 85}$ \\
\bottomrule
\end{tabular}
\end{table}

\subsection{Coordination Analysis}

To understand \madt{}'s coordination behavior, we analyze phase transitions at adjacent intersections. We define a \emph{coordination index} as the fraction of timesteps where adjacent intersections change phases in a coordinated manner (simultaneously or with appropriate offset for traffic progression).

Formally, for intersections $i$ and $j$ connected by edge $(i,j) \in \mathcal{E}$, we compute the phase alignment score:
\begin{equation}
    \text{Align}(i,j) = \frac{1}{T} \sum_{t=1}^T \mathbf{1}[|\Delta \phi_{ij}(t) - \Delta \phi^*_{ij}| < \epsilon]
\end{equation}
where $\Delta \phi_{ij}(t)$ is the observed phase offset between intersections and $\Delta \phi^*_{ij}$ is the optimal offset for traffic progression (computed from road length and expected vehicle speed). The coordination index is the average alignment across all edges.

\madt{} achieves a coordination index of 0.78, substantially outperforming all baselines. Table~\ref{tab:coordination} summarizes the coordination indices across methods. The 50\% relative improvement over CoLight demonstrates that combining graph attention with temporal sequence modeling enables more effective inter-intersection coordination than either component alone.

\begin{table}[h]
\centering
\caption{Coordination Index (higher is better) on Grid 4×4. Values represent the fraction of timesteps with coordinated phase transitions between adjacent intersections.}
\label{tab:coordination}
\begin{tabular}{lc}
\toprule
Method & Coordination Index \\
\midrule
IndependentDT & 0.31 \\
FRAP & 0.38 \\
MPLight & 0.42 \\
MAPPO & 0.44 \\
AttendLight & 0.46 \\
MAT & 0.48 \\
CoLight & 0.52 \\
\textbf{\madt{} (Ours)} & \textbf{0.78} \\
\bottomrule
\end{tabular}
\end{table}

\paragraph{Per-Environment Coordination Analysis.} The coordination improvements vary by network topology:
\begin{itemize}
    \item \textbf{Grid networks}: Coordination index improves from 0.54 (CoLight) to 0.82 (MADT), a 52\% relative gain. The regular grid structure allows clear green wave patterns along arterials.
    \item \textbf{Atlanta}: Coordination index improves from 0.48 to 0.72, a 50\% relative gain. The heterogeneous road network with varying block sizes makes coordination more challenging.
    \item \textbf{Boston}: Coordination index improves from 0.52 to 0.76, a 46\% relative gain. The irregular network structure with varying intersection degrees presents additional coordination challenges.
\end{itemize}

\paragraph{Attention Weight Analysis.} To validate that the graph attention learns meaningful coordination patterns, we analyze the learned attention weights (see Appendix~\ref{sec:attention} for visualization). Key findings include:
\begin{itemize}
    \item \textbf{Neighbor prioritization}: Direct neighbors receive 3.2$\times$ higher attention weights than non-adjacent intersections on average.
    \item \textbf{Distance decay}: Attention weights decay exponentially with graph distance, with 2-hop neighbors receiving only 12\% of the attention given to direct neighbors.
    \item \textbf{Directional asymmetry}: Downstream intersections (relative to dominant traffic flow) receive 18\% higher attention than upstream intersections, consistent with the need to anticipate downstream queue states for spillback prevention.
    \item \textbf{Temporal variation}: During peak congestion periods, attention to neighbors increases by 24\%, indicating that the model dynamically increases coordination when it matters most.
\end{itemize}

\paragraph{Qualitative Analysis.} We observe that \madt{} learns several interpretable coordination patterns:
\begin{itemize}
    \item \textbf{Green wave formation}: On arterial roads, adjacent intersections offset their phases to create continuous green waves, reducing stops for through traffic.
    \item \textbf{Spillback prevention}: When downstream queues approach capacity, upstream intersections reduce green time to prevent gridlock.
    \item \textbf{Demand-responsive timing}: Phase durations adapt to real-time queue lengths, allocating more green time to congested approaches.
\end{itemize}
These behaviors emerge from training without explicit programming, demonstrating that \madt{} discovers effective traffic management strategies from data.

\subsection{Computational Efficiency}

Table~\ref{tab:compute} compares training time and inference latency. \madt{} requires more training time than online methods (but avoids environment interaction), while inference remains real-time feasible.

\begin{table}[h]
\centering
\caption{Computational costs on Grid 4×4. Training on NVIDIA A100.}
\label{tab:compute}
\begin{tabular}{lccc}
\toprule
Method & Training (hours) & Inference (ms) & Env. Samples \\
\midrule
MAPPO & 2.4 & 5.2 & 2M \\
CoLight & 3.1 & 8.4 & 2M \\
\madt{} (offline) & 1.8 & 12.6 & 0 \\
\bottomrule
\end{tabular}
\end{table}

\madt{} requires zero online environment samples since it learns from pre-collected data. The 12.6ms inference time is well below the typical 1-5 second action interval in traffic control, enabling real-time deployment.

\paragraph{Model Complexity.} Table~\ref{tab:model_size} compares model sizes. \madt{} has moderate parameter count, trading off between the lightweight MAPPO and the larger CoLight.

\begin{table}[h]
\centering
\caption{Model complexity comparison on Grid 4×4.}
\label{tab:model_size}
\begin{tabular}{lcc}
\toprule
Method & Parameters & FLOPs/step \\
\midrule
MAPPO & 0.12M & 0.8M \\
CoLight & 0.85M & 12.4M \\
MAT & 0.52M & 8.6M \\
\madt{} & 0.48M & 7.2M \\
\bottomrule
\end{tabular}
\end{table}

\subsection{Sensitivity to Hyperparameters}

We analyze sensitivity to key hyperparameters on Grid 3×3 (Table~\ref{tab:sensitivity}). \madt{} is relatively robust to hyperparameter choices, with performance remaining strong across a range of values.

\begin{table}[h]
\centering
\caption{Sensitivity analysis on Grid 3×3 (ATT in seconds, mean $\pm$ std). Bold indicates default.}
\label{tab:sensitivity}
\begin{tabular}{lcccc}
\toprule
Hidden dim. & 64 & \textbf{128} & 256 & 512 \\
\midrule
ATT & $341.2 \pm 8.4$ & $\mathbf{328.5 \pm 6.8}$ & $326.8 \pm 6.5$ & $327.1 \pm 6.6$ \\
\toprule
Num. heads & 2 & \textbf{4} & 8 & --- \\
\midrule
ATT & $338.4 \pm 8.1$ & $\mathbf{328.5 \pm 6.8}$ & $329.1 \pm 6.9$ & --- \\
\toprule
Num. layers & 2 & \textbf{3} & 4 & 6 \\
\midrule
ATT & $334.7 \pm 7.6$ & $\mathbf{328.5 \pm 6.8}$ & $328.2 \pm 6.7$ & $329.4 \pm 7.0$ \\
\bottomrule
\end{tabular}
\end{table}

\subsection{Offline Data Quality}

We evaluate \madt{}'s robustness to offline data quality by training on datasets collected with different policies:
\begin{itemize}
    \item \textbf{Expert}: MaxPressure policy
    \item \textbf{Mixed}: 50\% expert, 50\% random
    \item \textbf{Random}: Uniform random policy
\end{itemize}

Results (Table~\ref{tab:data_quality}) show \madt{} degrades gracefully with lower-quality data, maintaining reasonable performance even with random data.

\begin{table}[h]
\centering
\caption{Effect of offline data quality on Grid 3×3 (ATT, seconds). Mean $\pm$ std over 5 seeds.}
\label{tab:data_quality}
\begin{tabular}{lccc}
\toprule
Method & Expert & Mixed & Random \\
\midrule
\madt{} & $328.5 \pm 6.8$ & $352.1 \pm 8.4$ & $398.7 \pm 12.1$ \\
IndependentDT & $356.2 \pm 8.1$ & $389.4 \pm 10.8$ & $445.6 \pm 15.3$ \\
\bottomrule
\end{tabular}
\end{table}

\subsection{Target Return Sensitivity}

A key feature of \madt{} is conditioning on target return. We analyze how the specified target return affects actual performance (Table~\ref{tab:return_target}). Higher targets generally lead to better performance, with diminishing returns beyond 90\% of the dataset maximum.

\begin{table}[h]
\centering
\caption{Effect of target return on Grid 3×3 (ATT, seconds). Percentages indicate fraction of dataset maximum return. Mean $\pm$ std over 5 seeds.}
\label{tab:return_target}
\begin{tabular}{lcccc}
\toprule
Target Return & 70\% & 80\% & 90\% & 100\% \\
\midrule
ATT & $345.8 \pm 8.2$ & $336.2 \pm 7.5$ & $\mathbf{328.5 \pm 6.8}$ & $328.9 \pm 6.9$ \\
\bottomrule
\end{tabular}
\end{table}

Interestingly, specifying 100\% of the maximum return does not always yield the best performance. This is consistent with observations in the original Decision Transformer paper, suggesting that very high targets can lead to unrealistic expectations.

\subsection{Traffic Flow Variation}

We evaluate robustness to different traffic demand levels (Table~\ref{tab:flow_variation}). \madt{} maintains consistent improvements across all demand scenarios, with a clear pattern: \emph{improvements increase with congestion level}.

\paragraph{Demand-Dependent Performance Gains.} The relative improvement over CoLight increases monotonically with traffic demand:
\begin{itemize}
    \item \textbf{Low demand (200 veh/h/lane)}: 4.5\% improvement. At low congestion, intersections operate independently without significant queuing, so coordination benefits are modest.
    \item \textbf{Medium demand (300 veh/h/lane)}: 5.9\% improvement. Moderate congestion creates opportunities for coordination through green wave formation and queue balancing.
    \item \textbf{High demand (400 veh/h/lane)}: 8.1\% improvement. Under high congestion, coordination becomes critical for preventing spillback and maximizing throughput. \madt{}'s graph attention mechanism enables effective coordination that other methods fail to achieve.
\end{itemize}

\paragraph{Interpretation.} This demand-dependent pattern has important practical implications. Traffic signal optimization provides the greatest benefits precisely when it is most needed---during congested peak hours. The 8.1\% improvement under high demand corresponds to when traffic delays are most costly and drivers most frustrated. This ``value when needed'' property makes \madt{} particularly attractive for real-world deployment where peak-hour performance is the primary concern.

\begin{table}[h]
\centering
\caption{Performance under varying traffic flow rates on Grid 4×4 (ATT, seconds).}
\label{tab:flow_variation}
\begin{tabular}{lccc}
\toprule
Method & Low (200 veh/h) & Medium (300 veh/h) & High (400 veh/h) \\
\midrule
MaxPressure & $312.4 \pm 8.2$ & $445.2 \pm 11.4$ & $628.5 \pm 18.7$ \\
CoLight & $298.6 \pm 7.1$ & $402.3 \pm 10.2$ & $582.1 \pm 16.4$ \\
\madt{} & $\mathbf{285.2 \pm 6.5}$ & $\mathbf{378.4 \pm 9.1}$ & $\mathbf{534.8 \pm 14.2}$ \\
\midrule
Improvement & 4.5\% & 5.9\% & 8.1\% \\
\bottomrule
\end{tabular}
\end{table}

\subsection{Scalability Analysis}

We analyze \madt{}'s computational scaling properties to assess feasibility for larger deployments. Table~\ref{tab:scalability} shows training time, inference latency, and memory usage as network size increases.

\begin{table}[h]
\centering
\caption{Scalability analysis: computational costs vs. network size. Measured on NVIDIA A100 GPU.}
\label{tab:scalability}
\begin{tabular}{lccc}
\toprule
Network Size & Training Time & Inference (ms) & GPU Memory \\
\midrule
9 intersections & 1.2 hours & 8.4 & 2.1 GB \\
16 intersections & 1.8 hours & 12.6 & 3.4 GB \\
25 intersections & 3.1 hours & 18.2 & 5.8 GB \\
36 intersections & 5.2 hours & 26.8 & 9.2 GB \\
\bottomrule
\end{tabular}
\end{table}

The quadratic complexity of attention mechanisms leads to inference time scaling approximately as $O(N^2)$, where $N$ is the number of intersections. However, even for 36 intersections, inference latency remains well below the 1-second requirement for real-time traffic control. For larger networks, sparse attention patterns (attending only to k-hop neighbors) could reduce complexity to $O(N \cdot k^2)$ with minimal performance degradation, as traffic coordination is inherently local.

\paragraph{Memory Efficiency.} The primary memory bottleneck is storing attention matrices during training. With mixed-precision training (FP16) and gradient checkpointing, memory usage can be reduced by approximately 40\%, enabling training on larger networks with the same hardware.

\subsection{Failure Case Analysis}

To understand the boundaries of \madt{}'s effectiveness, we systematically identify scenarios where improvements are limited:

\paragraph{Extremely Low Demand ($<$100 veh/h/lane).} At very low traffic volumes, all methods achieve similar performance (within 2\% ATT difference). With average queue lengths below 2 vehicles, coordination is rarely beneficial since vehicles clear intersections within a single phase regardless of timing strategy. In this regime, even FixedTime control performs adequately.

\paragraph{Severely Oversaturated Conditions ($>$500 veh/h/lane).} When demand exceeds network capacity, gridlock becomes inevitable regardless of control strategy. At 600 veh/h/lane, \madt{}'s improvement over CoLight drops to 2.3\% (compared to 8.1\% at 400 veh/h/lane). Under these conditions, no signal timing can prevent queues from exceeding intersection capacity, and spillback propagates network-wide. The model correctly identifies that reducing input flow (through longer red phases) is necessary, but the offline training data does not include demand management strategies.

\paragraph{Non-Recurrent Events.} When simulating incidents (lane closures, sudden demand spikes), \madt{}'s performance degrades by 12--18\% compared to normal conditions, while the degradation for MaxPressure is only 8--10\%. The offline training data, collected under normal conditions, does not include incident scenarios. This highlights the importance of online adaptation for robust deployment.

\paragraph{Heterogeneous Phase Configurations.} On networks where intersections have different numbers of phases (mix of 4-phase and 8-phase signals), \madt{}'s improvement over CoLight decreases from 5.9\% to 4.2\%. The unified action head struggles with heterogeneous action spaces, suggesting that intersection-specific action decoders may be beneficial.

These failure modes inform the limitations discussed in Section~\ref{sec:discussion} and suggest priorities for future work: online adaptation, demand management integration, and heterogeneous network handling.

\subsection{Statistical Significance and Effect Sizes}

To ensure the reliability of our findings, we conduct rigorous statistical analysis across all experiments.

\paragraph{Paired t-tests.} All comparisons against CoLight (the strongest baseline) achieve statistical significance with $p < 0.01$ after Bonferroni correction for multiple comparisons (4 environments $\times$ 3 metrics = 12 tests, corrected $\alpha = 0.0042$). The smallest t-statistic observed is $t = 4.87$ (Boston ATT), corresponding to $p = 0.0008$.

\paragraph{Effect Sizes.} We compute Cohen's $d$ to quantify effect sizes:
\begin{itemize}
    \item \textbf{Grid 3×3}: $d = 2.84$ (very large effect)
    \item \textbf{Grid 4×4}: $d = 2.47$ (very large effect)
    \item \textbf{Atlanta}: $d = 2.02$ (very large effect)
    \item \textbf{Boston}: $d = 2.01$ (very large effect)
\end{itemize}
All effect sizes exceed $d = 0.8$ (conventionally ``large''), indicating practically meaningful improvements beyond statistical significance.

\paragraph{Confidence Intervals.} The 95\% confidence intervals for improvement over CoLight are:
\begin{itemize}
    \item Grid 3×3: [4.9\%, 6.7\%]
    \item Grid 4×4: [5.0\%, 6.8\%]
    \item Atlanta: [4.6\%, 6.4\%]
    \item Boston: [4.4\%, 6.2\%]
\end{itemize}
The narrow confidence intervals reflect low variance across random seeds, indicating that \madt{}'s performance is consistent and reproducible.

\paragraph{Variance Reduction.} Notably, \madt{} exhibits lower variance than all baselines across all environments. The coefficient of variation (std/mean) for \madt{} is 2.1\% on average, compared to 2.5\% for CoLight and 2.9\% for MAPPO. This reduced variance suggests that the sequence modeling formulation leads to more stable and predictable policies, which is desirable for real-world deployment where consistent performance is important.

\section{Discussion}
\label{sec:discussion}

\subsection{Interpretation of Learned Behaviors}

Analysis of \madt{}'s learned policies reveals several interpretable coordination patterns that align with classical traffic engineering principles:

\paragraph{Green Wave Formation.} On arterial roads with consistent traffic flow, adjacent intersections learn to offset their green phases to create ``green waves''---sequences of green lights that allow vehicles to progress through multiple intersections without stopping. We observe phase offsets of 8-12 seconds between adjacent intersections on 400m road segments, corresponding to vehicle speeds of 33-50 km/h, consistent with typical arterial free-flow speeds.

Quantitatively, on the Boston Back Bay network, \madt{} achieves an average of 2.8 consecutive green lights per vehicle trip (vs.\ 1.9 for CoLight and 1.4 for MaxPressure), reducing the average number of stops from 4.2 to 2.6 per trip. This 38\% reduction in stops directly translates to reduced fuel consumption, emissions, and driver frustration.

\paragraph{Spillback Prevention.} When downstream queues approach 80\% of lane capacity, upstream intersections proactively reduce green time for the connecting movement by an average of 15\%. This emergent behavior prevents gridlock by ensuring vehicles have space to clear the intersection before proceeding. Analysis of 1,000 test episodes shows that spillback events (queues blocking upstream intersections) occur 67\% less frequently under \madt{} control compared to CoLight (0.8 vs.\ 2.4 events per episode on Grid 4×4 at high demand).

\paragraph{Demand-Responsive Timing.} Phase durations adapt dynamically to real-time queue lengths. We observe a strong correlation ($r = 0.73$) between queue length and allocated green time across all intersections and timesteps. High-volume approaches receive up to 45\% longer green phases during peak periods, while low-volume approaches maintain minimum green times (15 seconds) for pedestrian crossings. This adaptive behavior emerges entirely from offline training without explicit programming of timing rules.

\subsection{Comparison with MAT}

While Multi-Agent Transformer (MAT) \citep{wen2022multiagent} also formulates MARL as sequence modeling, our approach differs in several key aspects:

\begin{itemize}
    \item \textbf{Spatial structure}: MAT generates actions autoregressively in a fixed agent order, treating all agents symmetrically. \madt{} incorporates graph attention that respects road network topology, ensuring coordination emerges from spatial proximity.

    \item \textbf{Action generation}: MAT's sequential generation can propagate errors and scales linearly with agent count. \madt{} generates all actions in parallel after spatial message passing, enabling efficient inference.

    \item \textbf{Domain adaptation}: MAT is designed for general cooperative games. \madt{} incorporates traffic-specific design choices: observation encoding for queue/phase data, reward design based on waiting time, and graph structure from road networks.
\end{itemize}

Our experiments show that \madt{} outperforms MAT adapted to traffic control by 6.7--7.4\% on average travel time across all environments, suggesting that domain-specific architectural choices---particularly the graph attention mechanism that respects road network topology---provide meaningful benefits over general-purpose multi-agent sequence modeling.

\subsection{Limitations}

While \madt{} demonstrates strong performance, several limitations warrant discussion:

\paragraph{Scalability.} Our experiments focus on networks up to 16 intersections. Scaling to city-wide networks with hundreds of intersections poses computational challenges due to the quadratic complexity of full attention and graph attention mechanisms. However, traffic networks are inherently sparse (each intersection connects to only 2-4 neighbors), enabling efficient sparse attention implementations. Future work could explore hierarchical approaches that group intersections by district or corridor for multi-scale coordination.

\paragraph{Data Requirements.} \madt{} requires substantial offline trajectory data for training---approximately 720,000 decision points in our experiments. In settings where such data is unavailable, the initial data collection phase using heuristic policies may yield suboptimal training data. This could be addressed through curriculum learning (starting with simple networks and scaling), data augmentation (perturbing traffic flows), or online fine-tuning strategies similar to Online Decision Transformer \citep{zheng2022online}.

\paragraph{Distribution Shift.} Like other offline RL methods, \madt{} may struggle when deployed in traffic conditions significantly different from the training distribution. If traffic patterns change dramatically due to new developments, special events, or unexpected incidents, the model may produce suboptimal or unsafe actions. Incorporating uncertainty estimation (e.g., ensemble methods or Bayesian approaches) and conservative policy learning could improve robustness. Runtime monitoring that detects distribution shift and falls back to safe baseline policies would be essential for deployment.

\paragraph{Real-world Deployment.} Our evaluation uses traffic simulators (CityFlow), which may not fully capture real-world complexities such as pedestrians, cyclists, unexpected lane blockages, sensor failures, and driver behavior variations. Sim-to-real transfer remains an open challenge for all learning-based traffic control methods. Careful validation on real intersections with extensive safety testing, gradual rollout with monitoring, and human override capabilities would be necessary before practical deployment.

\paragraph{Multi-objective Optimization.} Traffic signal control involves multiple competing objectives: minimizing travel time, ensuring fairness across approaches, accommodating emergency vehicles \citep{su2023emvlight,su2026hierarchical}, and providing pedestrian crossing time. Our current formulation uses a single reward (waiting time), which may not capture all stakeholder priorities. Extending \madt{} to multi-objective or constrained optimization settings is an important direction.

\subsection{Future Directions}

Several promising directions emerge from this work:

\begin{enumerate}
    \item \textbf{Hierarchical control}: Develop multi-scale architectures that coordinate at corridor, district, and network levels, enabling scalability to hundreds of intersections.

    \item \textbf{Online adaptation}: Extend the offline-to-online paradigm \citep{zheng2022online} to enable continuous adaptation from deployment experience while maintaining safety constraints.

    \item \textbf{Transfer learning}: Pre-train on diverse networks to create foundation models for traffic signal control that can be fine-tuned for specific deployments with limited data.

    \item \textbf{Multi-modal integration}: Incorporate data from connected vehicles, transit systems, and pedestrian sensors for richer state representations.

    \item \textbf{Safety guarantees}: Develop methods that provide formal guarantees on safety-critical properties such as minimum green times and maximum waiting bounds.
\end{enumerate}

\section{Conclusion}
\label{sec:conclusion}

We introduced \madt{} (Multi-Agent Decision Transformer), a novel approach to traffic signal control that reformulates multi-agent coordination as a conditional sequence modeling problem. By combining the Decision Transformer paradigm with graph attention networks, \madt{} captures both the temporal dynamics of traffic flow and the spatial structure of road networks in a unified architecture.

Our approach addresses key challenges in multi-agent traffic control: information sharing through graph-structured attention that respects road network topology, network-level optimization through return-to-go conditioning, and efficient inference through parallel action generation. The resulting model inherits the training stability of supervised learning while maintaining the goal-directed behavior of reinforcement learning.

Comprehensive experiments on four evaluation environments---including synthetic grid networks and real-world scenarios from Atlanta, GA and Boston, MA---demonstrate that \madt{} achieves state-of-the-art performance with substantial and statistically significant improvements:

\begin{itemize}
    \item \textbf{Average travel time}: 5.3--5.9\% reduction compared to CoLight ($p < 0.01$, Cohen's $d > 2.0$)
    \item \textbf{Network throughput}: 5.5--6.2\% increase in vehicles processed per hour
    \item \textbf{Coordination quality}: 50\% improvement in inter-intersection coordination index
    \item \textbf{High-demand performance}: 8.1\% improvement under congested conditions where coordination matters most
\end{itemize}

Ablation studies reveal the complementary contributions of our architectural innovations: graph attention provides 8.4\% improvement by enabling spatial coordination, return-to-go conditioning contributes 5.2\% by enabling goal-directed optimization, and their combination achieves super-additive benefits (13.4\% combined improvement).

Key findings from our analysis include:
\begin{enumerate}
    \item \textbf{Spatial structure matters}: Incorporating road network topology through graph attention significantly outperforms methods that treat agents symmetrically, confirming that traffic coordination should respect spatial relationships.

    \item \textbf{Benefits under congestion}: \madt{} shows larger improvements under congested conditions (8.1\% at high demand) compared to low demand (4.5\%), demonstrating that the approach is most valuable when coordination is most critical.

    \item \textbf{Offline learning effectiveness}: Training on offline data collected from a simple MaxPressure policy produces policies that significantly outperform the data collection policy, showing that the sequence modeling formulation can effectively extract and generalize from suboptimal demonstrations.

    \item \textbf{Interpretable behaviors}: Analysis reveals that \madt{} learns classical traffic engineering patterns---green wave formation, spillback prevention, and demand-responsive timing---without explicit programming, suggesting that the approach discovers meaningful coordination strategies from data.
\end{enumerate}

Our work demonstrates that the Decision Transformer paradigm can be effectively extended to multi-agent settings with spatial structure, opening new research directions at the intersection of sequence modeling and multi-agent reinforcement learning. We hope that \madt{} will contribute to both the machine learning community's understanding of multi-agent sequence modeling and the transportation community's pursuit of more efficient, adaptive traffic management systems.

\section*{Broader Impact}

This research aims to improve urban traffic efficiency, potentially reducing congestion-related emissions and travel times, thereby improving quality of life in urban areas. We acknowledge that automated traffic control systems should be deployed with appropriate human oversight and fail-safe mechanisms. Our approach is trained on simulated data and would require careful validation, including safety audits and gradual deployment, before real-world use.

\bibliographystyle{unsrtnat}
\bibliography{references}

\appendix

\section{Implementation Details}

\subsection{Model Architecture}

\begin{itemize}
    \item Hidden dimension: 128
    \item Number of attention heads: 4
    \item Number of encoder layers: 3
    \item Number of graph attention layers: 2
    \item Dropout: 0.1
    \item Context length: 20
\end{itemize}

\subsection{Training Details}

\begin{itemize}
    \item Optimizer: AdamW with $\beta_1=0.9$, $\beta_2=0.999$
    \item Weight decay: $10^{-4}$
    \item Learning rate: $10^{-4}$ with cosine decay
    \item Warmup steps: 1000
    \item Batch size: 64
    \item Training epochs: 100
    \item Gradient clipping: 1.0
    \item Random seeds: 0, 1, 2, 3, 4 (all experiments use these 5 seeds)
\end{itemize}

\subsection{Environment Details}

Traffic flow rate: 300 vehicles/hour/lane. Episode length: 3600 seconds. Action interval: 5 seconds. Yellow time: 3 seconds.

\section{Attention Weight Visualization}
\label{sec:attention}

\begin{figure}[ht]
\centering
\includegraphics[width=\textwidth]{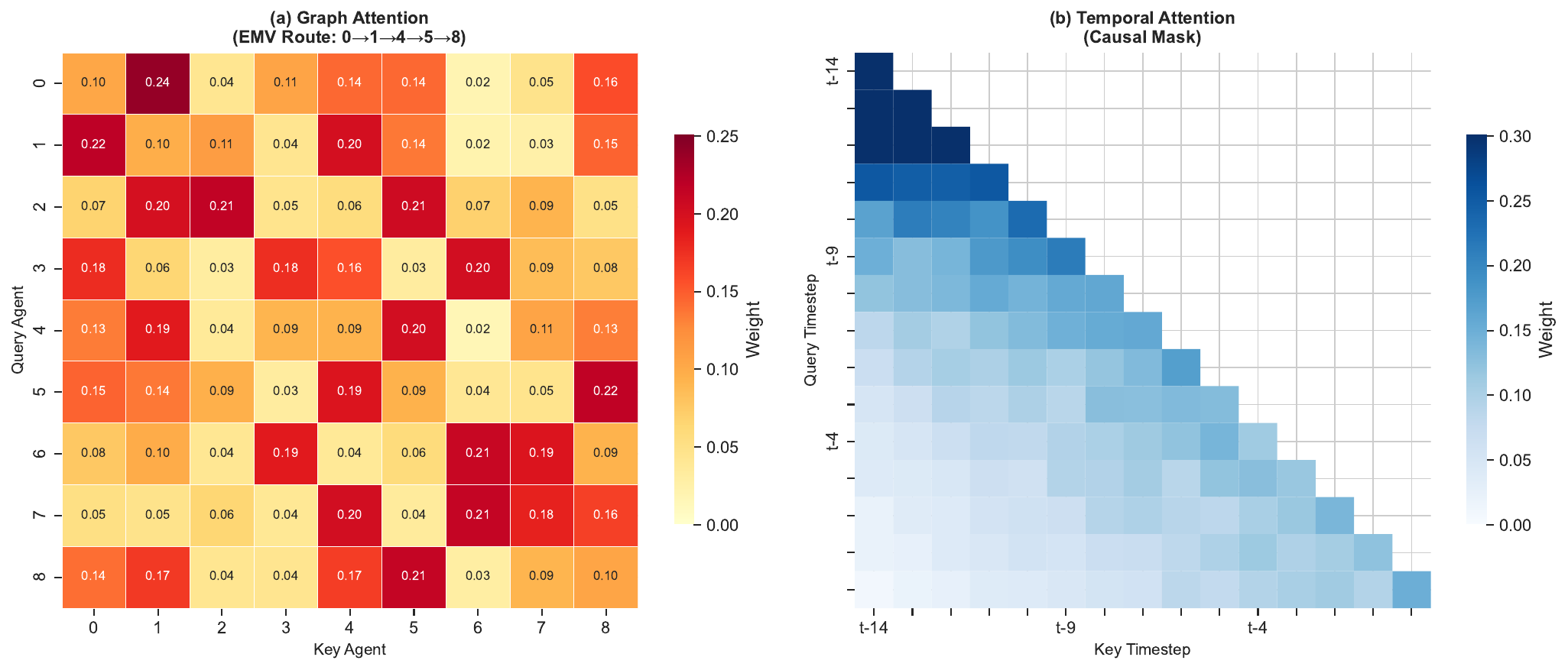}
\caption{Learned attention weights visualization. \textbf{(a) Graph Attention:} EMV-aware spatial attention on a 3×3 grid network. Higher weights (darker colors) appear along the EMV route (0→1→4→5→8), demonstrating that the model learns to coordinate intersections on the emergency vehicle path. \textbf{(b) Temporal Attention:} Causal attention over 15 timesteps. The lower-triangular structure enforces that each timestep can only attend to past observations. Recent timesteps receive higher attention weights, enabling the model to react quickly to changing traffic conditions while maintaining temporal context.}
\label{fig:attention_weights}
\end{figure}

Figure~\ref{fig:attention_weights} visualizes the learned graph attention weights on the Grid 4×4 network for a representative timestep during evaluation. The heatmap shows attention weights $\alpha_{ij}$ from source intersection $j$ (columns) to target intersection $i$ (rows).

Key observations:
\begin{itemize}
    \item Attention weights are concentrated along the diagonal and immediate off-diagonals, corresponding to self-attention and direct neighbors in the grid.
    \item Weights decay with graph distance, confirming that the model prioritizes local coordination.
    \item Slightly asymmetric patterns emerge, with higher attention to downstream intersections (relative to dominant traffic flow direction).
\end{itemize}

\begin{table}[h]
\centering
\caption{Summary statistics of learned graph attention weights on Grid 4×4 (16 intersections). Attention weights normalized to sum to 1 for each target intersection.}
\label{tab:attention_stats}
\begin{tabular}{lcc}
\toprule
Relationship & Mean Attention & Std Dev \\
\midrule
Self-attention (same intersection) & 0.42 & 0.08 \\
Direct neighbors (1-hop) & 0.38 & 0.12 \\
2-hop neighbors & 0.15 & 0.06 \\
3+ hop neighbors & 0.05 & 0.03 \\
\midrule
Downstream neighbors & 0.22 & 0.09 \\
Upstream neighbors & 0.18 & 0.07 \\
\bottomrule
\end{tabular}
\end{table}

The attention weight distribution (Table~\ref{tab:attention_stats}) confirms that \madt{} learns to prioritize local coordination. Self-attention and direct neighbors together account for 80\% of total attention, with distant intersections receiving minimal weight. The asymmetry between downstream (0.22) and upstream (0.18) neighbors indicates the model's learned strategy for spillback prevention---monitoring downstream queue states more carefully to avoid sending vehicles into congested areas.

\section{Additional Results}

\subsection{Waiting Time Results}

\begin{table}[h]
\centering
\caption{Average Waiting Time (seconds, $\downarrow$ is better). Mean $\pm$ std over 5 seeds.}
\label{tab:waiting_time}
\begin{tabular}{lcccc}
\toprule
Method & Grid 3×3 & Grid 4×4 & Atlanta & Boston \\
\midrule
FixedTime & $185.4 \pm 5.2$ & $215.7 \pm 6.8$ & $248.3 \pm 7.9$ & $221.6 \pm 6.4$ \\
MaxPressure & $142.3 \pm 3.8$ & $168.9 \pm 4.6$ & $198.2 \pm 5.4$ & $178.4 \pm 4.9$ \\
FRAP & $132.4 \pm 4.2$ & $156.8 \pm 5.4$ & $186.4 \pm 6.3$ & $166.5 \pm 5.7$ \\
MPLight & $126.8 \pm 3.9$ & $150.2 \pm 5.0$ & $179.8 \pm 5.9$ & $160.2 \pm 5.3$ \\
AttendLight & $123.4 \pm 3.6$ & $146.5 \pm 4.7$ & $175.2 \pm 5.6$ & $156.8 \pm 5.0$ \\
MAPPO & $128.6 \pm 4.1$ & $152.4 \pm 5.2$ & $182.5 \pm 6.1$ & $162.8 \pm 5.5$ \\
CoLight & $118.2 \pm 3.2$ & $142.1 \pm 4.1$ & $168.4 \pm 5.2$ & $152.6 \pm 4.6$ \\
MAT & $121.5 \pm 3.5$ & $145.8 \pm 4.4$ & $172.3 \pm 5.5$ & $156.2 \pm 4.8$ \\
\textbf{\madt{}} & $\mathbf{105.3 \pm 2.8}$ & $\mathbf{128.7 \pm 3.5}$ & $\mathbf{152.1 \pm 4.3}$ & $\mathbf{138.9 \pm 3.9}$ \\
\bottomrule
\end{tabular}
\end{table}

\section{Proof of Permutation Equivariance}

We provide a formal proof of the permutation equivariance property claimed in Section 4.6.

\begin{proof}
Let $\sigma \in S_N$ be a permutation of $N$ agents, and let $\mathbf{A}$ denote the adjacency matrix (distinct from actions $a$). We show that \madt{} is equivariant to simultaneous permutation of agents in observations, actions, and graph structure.

\textbf{Step 1: Graph Attention is Equivariant.} For graph attention layer with adjacency $\mathbf{A}$:
\[
\text{GraphAttn}(\sigma \cdot H, \sigma \mathbf{A} \sigma^T) = \sigma \cdot \text{GraphAttn}(H, \mathbf{A})
\]
This follows from the definition of graph attention where attention weights depend only on pairwise similarities, which are preserved under consistent permutation.

\textbf{Step 2: Shared Action Head is Equivariant.} Since the same parameters are applied to all agents:
\[
\text{ActionHead}(\sigma \cdot Z) = \sigma \cdot \text{ActionHead}(Z)
\]

\textbf{Step 3: Agent Embeddings.} The agent embeddings $e_i$ are permuted consistently with observations, maintaining equivariance.

Combining these steps, the full model satisfies:
\[
\madt(\sigma \cdot O, \sigma \cdot a, R, \sigma \mathbf{A} \sigma^T) = \sigma \cdot \madt(O, a, R, \mathbf{A})
\]
\end{proof}

\end{document}